\documentclass[journal]{IEEEtran}
\usepackage{graphicx}
\usepackage{booktabs}
\usepackage{multirow} 
\usepackage{tabularx} 
\usepackage{adjustbox}
\usepackage{color, colortbl}
\definecolor{gray}{gray}{0.85}
\usepackage{nicematrix,tikz}
\usepackage{lipsum}

\usepackage{algorithm}
\usepackage{setspace}

\usepackage{algpseudocode}
\usepackage{verbatim}
\usepackage{amssymb}
\usepackage{makecell}
\usepackage{tikz}
\usepackage{hyperref}
\hypersetup{
    colorlinks=true,    
    linkcolor=blue,     
    filecolor=magenta,  
    urlcolor=cyan,      
    citecolor=blue,     
    pdfborder={0 0 0}   
}

\newcolumntype{Y}{>{\centering\arraybackslash}X}

\ifCLASSINFOpdf
\else
\fi
\begin{document}

%
\title{Collaborative Static-Dynamic Teaching: \\ A Semi-Supervised Framework for Stripe-Like Space Target Detection}
%
%
%

\author{Zijian~Zhu,
        Ali~Zia, \textit{Member, IEEE},
        Xuesong~Li, \textit{Member, IEEE},
        Bingbing~Dan, \\
        Yuebo~Ma,
        Hongfeng~Long, 
        Kaili~Lu,
        Enhai~Liu,
        and~Rujin~Zhao
\thanks{Manuscript received xx xx, 2024; revised xx xx, 2024; accepted xx xx, 2024. This work was supported in part by China Scholarship Council No. 202304910543, the Sichuan Outstanding Youth Science and Technology Talent Project No. 2022JDJQ0027, and the Department of Science and Technology of Sichuan Province No. 2022JDRC0065 and No. 2024NSFSC1443. (Corresponding authors:  Rujin Zhao; Ali Zia.)}
\thanks{Zijian~Zhu, Bingbing~Dan, Yuebo~Ma, Hongfeng~Long, Kaili~Lu, Enhai~Liu, and~Rujin~Zhao are with the National Key Laboratory of Optical Field Manipulation Science and Technology, Chinese Academy of Sciences, Chengdu 610209, China, also with the Institute of Optics and Electronics, Chinese Academy of Sciences, Chengdu 610209, China, also with the Key Laboratory of Science and Technology on Space Optoelectronic Precision Measurement, Chinese Academy of Sciences, Chengdu 610209, China, and also with the University of Chinese Academy of Sciences, Beijing 100049, China. (e-mail: zhuzijian20@mails.ucas.ac.cn; zhaorj@ioe.ac.cn;)}
\thanks{Ali~Zia and Xuesong~Li are with the College of Science, Australian National University,Canberra, 2601, Australia. }}

%
%

\markboth{IEEE TRANSACTIONS ON GEOSCIENCE AND REMOTE SENSING,~Vol.~xx, 2024}%
{Shell \MakeLowercase{\textit{et al.}}: Bare Demo of IEEEtran.cls for Journals}
%



\maketitle



\begin{abstract}

Stripe-like space target detection (SSTD) is crucial for space situational awareness. Traditional unsupervised methods often fail in low signal-to-noise ratio and variable stripe-like space targets scenarios, leading to weak generalization. Although fully supervised learning methods improve model generalization, they require extensive pixel-level labels for training. In the SSTD task, manually creating these labels is often inaccurate and labor-intensive. Semi-supervised learning (SSL) methods reduce the need for these labels and enhance model generalizability, but their performance is limited by pseudo-label quality. To address this, we introduce an innovative Collaborative Static-Dynamic Teacher (CSDT) SSL framework, which includes static and dynamic teacher models as well as a student model. This framework employs a customized adaptive pseudo-labeling (APL) strategy, transitioning from initial static teaching to adaptive collaborative teaching, guiding the student model's training. The exponential moving average (EMA) mechanism further enhances this process by feeding new stripe-like knowledge back to the dynamic teacher model through the student model, creating a positive feedback loop that continuously enhances the quality of pseudo-labels. Moreover, we present MSSA-Net, a novel SSTD network featuring a multi-scale dual-path convolution (MDPC) block and a feature map weighted attention (FMWA) block, designed to extract diverse stripe-like features within the CSDT SSL training framework. Extensive experiments verify the state-of-the-art performance of our framework on the AstroStripeSet and various ground-based and space-based real-world datasets. 

\end{abstract}

\section{Introduction}
\label{sec:intro}

\begin{figure}[ht]
  \centering
  \includegraphics[width=\columnwidth]{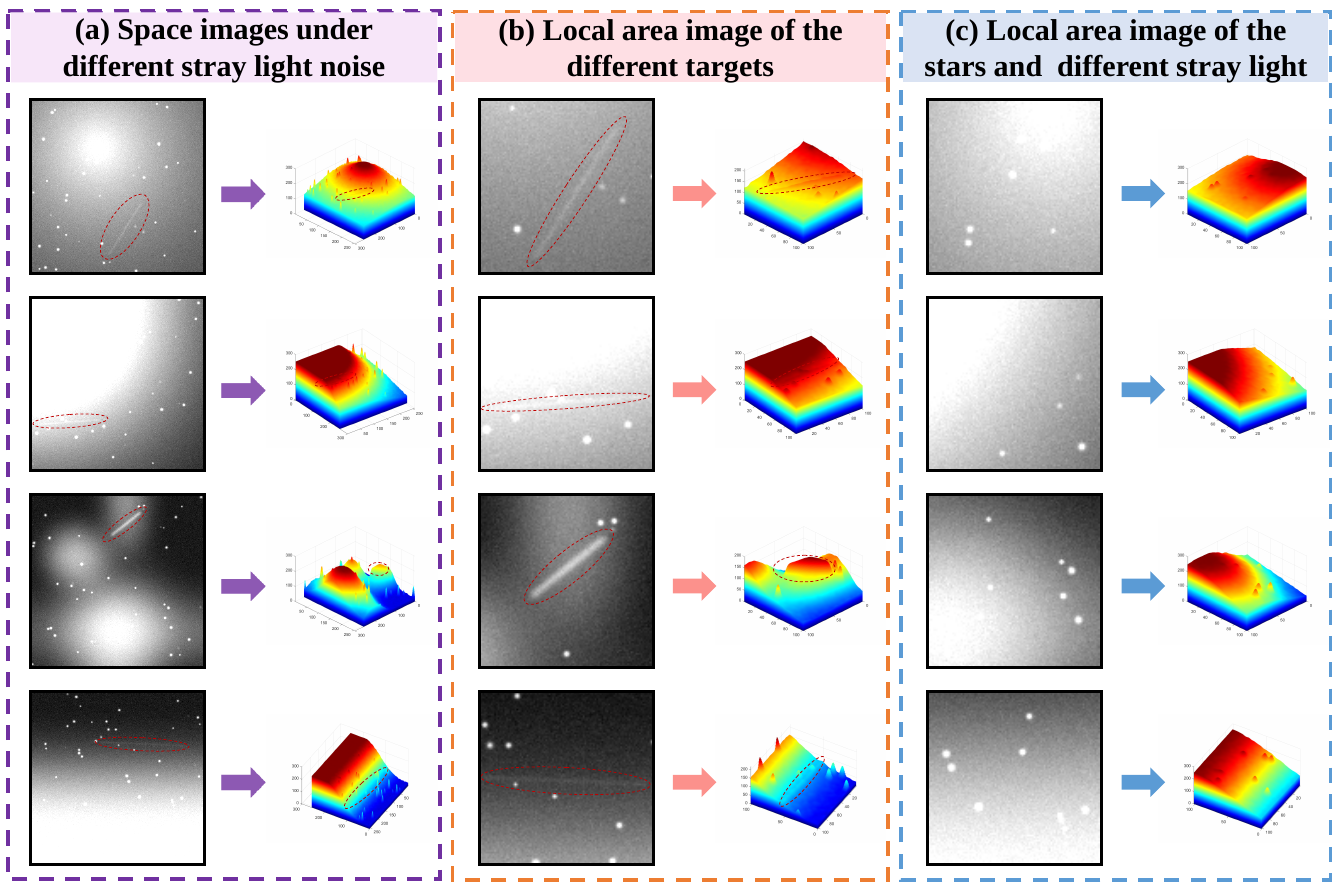}
  \caption{Representative space images with low signal-to-noise ratio stripe-like targets and their 3D pixel-level views under different stray light noise. (a) Space images affected by stray light and their 3D pixel-level views. (b) Detailed local images of stripe-like targets and their 3D pixel-level views. (c) Detailed local images of stars and stray light, along with their 3D pixel-level views. The stripe-like targets are highlighted with the dotted \textcolor{red}{red} circle.}
  \label{fig:fig1}
\end{figure}

\IEEEPARstart{T}{he} increasing number of permanent objects in space, such as satellites, spacecraft, and debris from human activities, makes space target detection an essential and ongoing task. \cite{li2020space, wirnsberger2015space,zhang2022dynamics}. Space target detection based on optical imaging systems has always been a hot research topic \cite{liu2020topological,diprima2018efficient,liu2020space1}. These optical imaging systems operate in two main configurations: target tracking and star tracking. These configurations correspond to three modes based on exposure time. In both configurations, short exposure times render targets and stars as point-like (mode 1) \cite{yao2022adaptive}. In star tracking with long exposures, targets appear as stripe-like and stars as point-like (mode 2) \cite{lin2021new,jiang2022automatic}. Conversely, in target tracking with long exposures, targets are point-like,  and stars appear as stripe-like (mode 3) \cite{felt2024seeing}. 
Each mode corresponds to a distinct research track, with this paper focusing on stripe-like space target detection (SSTD) in mode 2. Currently, the SSTD research faces two main challenges. As shown in Fig. \ref{fig:fig1}, the variability in the direction, length, and brightness of these stripe-like space targets pose the first challenge to the generalization capabilities of the SSTD methods. Moreover, long exposures enhance the visibility of faint and distant targets but also amplify background light, increasing stray light from reflections or scattering off celestial bodies \cite{lu2023fast,xu2020stray,li2022bsc,liu2023multi}. This increase in stray light lowers the signal-to-noise ratio (SNR), presenting a second challenge for the detection accuracy of existing SSTD methods. This underscores the need for advanced technologies to effectively address these challenges.

Current research on SSTD focuses mainly on traditional unsupervised methods \cite{diprima2018efficient,liu2020space1,jiang2022automatic,hickson2018fast,levesque2007image,levesque2009automatic} and convolutional neural networks (CNN) based methods \cite{li2022bsc,liu2023multi,jia2020detection}. 
Traditional unsupervised methods, which rely on manually customized filters or morphological operations, lack generalization capabilities in space scenarios with variable stripe-like targets and diverse low SNR conditions affected by stray light. Therefore, this lack of generalization capabilities has prompted a gradual shift towards CNN-based methods. However, existing CNN-based methods lack networks specifically designed for variable stripe-like targets, limiting their ability to effectively extract diverse stripe-like features during the learning process and reducing their effectiveness in unseen scenarios. 

To address this, we developed a multi-scale stripe attention network (MSSA-Net) specifically for SSTD task. This network extracts multi-scale stripe features using a tailored multi-scale dual-path convolution (MDPC) block to broaden the receptive field, and a feature map weighted attention (FMWA) block to enhance the feature map, improving sensitivity to the changing and low SNR stripe-like patterns.

\begin{figure}[ht]
  \centering
  \includegraphics[width=\columnwidth]{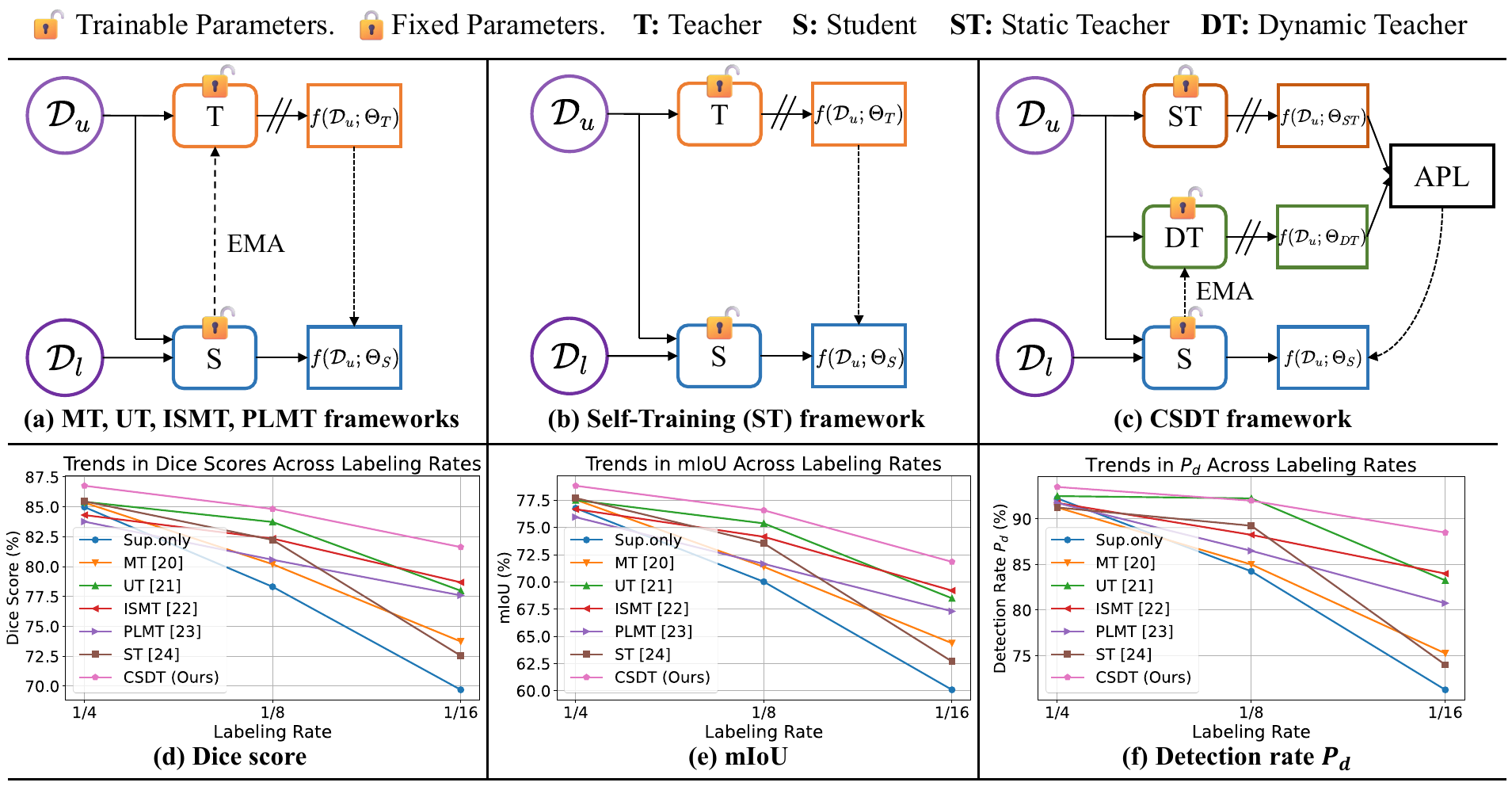}
  \caption{Comparison of SSL architectures with MSSA-Net as the teacher-student network: (a) Single-teacher architectures based on exponential moving average (EMA) strategies, including Mean Teacher (MT) \cite{tarvainen2017mean}, Unbiased Teacher (UT) \cite{liu2021unbiased}, Interactive Self-Training Mean Teacher (ISMT) \cite{yang2021interactive}, and Pseudo-Label Mean Teacher (PLMT) \cite{mao2023semi}. (b) Self-Training (ST) \cite{yang2022st}, a single-teacher architecture without EMA. (c) Our proposed Collaborative Static-Dynamic Teaching (CSDT) architecture utilizing EMA. Panels (d), (e), (f) show the performances of these SSL architectures in terms of Dice score, mIoU, and detection rate $P_d$ across various labeling rates.}

  \label{fig:fig2}
\end{figure}

On the other hand, existing CNN-based methods rely on a fully supervised learning (FSL) approach, which requires a large number of pixel-level labeled space images. Due to issues like stray light in space images, obtaining manual labels for stripe-like targets is often inaccurate and labor-intensive. Therefore, we turn our attention to semi-supervised learning (SSL), which reduces reliance on labeled space images. SSL enhances model generalization by using a large amount of unlabeled images to automatically discover new stripe-like patterns during training. As illustrated in Fig.~\hyperref[fig:fig2]{\ref*{fig:fig2}(a)} and Fig.~\hyperref[fig:fig2]{\ref*{fig:fig2}(b)}, the teacher-student architecture is the common framework in current SSL methods, with most employing a single teacher-student setup. However, this fixed single-teacher teaching mode leads to the teacher-student sub-network prematurely adapt to stripe-like pseudo-labels in space scenarios, causing co-training to effectively degenerate into self-training \cite{shen2023co}. This results in poor generalization under different stray light conditions.

To address this issue, we developed the collaborative static-dynamic teaching (CSDT) SSL framework, showcased in Fig.~\hyperref[fig:fig2]{\ref*{fig:fig2}(c)}. It employs a dual-teacher setup, consisting of a static teacher (ST) model, a dynamic teacher (DT) model, and a student (S) model. During training, the ST model is pre-trained with labeled images and remains fixed, while the DT model is updated iteratively through the exponential moving average (EMA) of the S model. Initially, the ST model provides stable guidance to the S model using its pre-trained knowledge. This is followed by a transition to adaptive collaborative teaching involving both ST and DT models, facilitated by a customized adaptive pseudo-labeling (APL) strategy, which introduces flexibility in the learning process. This dual-teacher setup allows the DT model to adapt to pseudo-labels more gradually, reducing the risk of overfitting compared to single-teacher methods. During inference, only the DT model is used, minimising additional time consumption.

As shown in Figs.~\hyperref[fig:fig2]{\ref*{fig:fig2}(d)--(f)}, we compare the SSTD performance of different SSL methods on AstroStripeSet \cite{zhu2024sstd} at three labeling rates, using MSSA-Net as the teacher-student network. The evaluation metrics include Dice score, mean intersection over union (mIoU), and detection rate ($P_d$). These results not only demonstrate the feasibility of SSL methods for the SSTD task but also highlight the significant advantages of the proposed CSDT framework. The compared methods include supervised only (Sup.only), Mean Teacher (MT) \cite{tarvainen2017mean}, Unbiased Teacher (UT) \cite{liu2021unbiased}, Interactive Self-Training Mean Teacher (ISMT) \cite{yang2021interactive}, and Pseudo-Label Mean Teacher (PLMT) \cite{mao2023semi}, as well as Self-Training (ST) \cite{yang2022st}.

Our contributions are summarised as follows:

\begin{itemize}
\item We propose CSDT, a novel SSL architecture that leverages unlabeled space images to learn stripe-like patterns, significantly enhancing model generalization. This approach reduces the reliance on extensive, inaccurate, and labor-intensive pixel-level labeled images, marking the first introduction of SSL techniques to the SSTD task.

\item We develop an APL strategy that identifies the optimal pseudo-labels and implement an EMA mechanism to dynamically enhances the performance of the DT model as the S model progresses, creating a beneficial feedback loop within the CSDT framework.

\item We design MSSA-Net specifically for SSTD, incorporating MDPC and FMWA blocks in the encoder stage to significantly enhance its responsiveness to diverse and faint stripe-like patterns.

\item Extensive experiments demonstrate that the proposed CSDT framework achieves state-of-the-art (SOTA) performance on the AstroStripeSet, and exhibit robust zero-shot generalization capabilities across various ground-based and space-based real-world image datasets.
\end{itemize}

The rest of this paper is is organised into several sections. Section \ref{sec:related work} reviews the related work. Section \ref{sec:Methodology} details our CSDT Framework for SSTD. Section \ref{sec:exper} presents experimental validation results. Section \ref{sec:conc} concludes our findings.

\section{Related Work}
\label{sec:related work}

This section reviews the related methods, focusing on traditional unsupervised approaches, FSL methods, and SSL techniques.

\subsection{Traditional Unsupervised Methods for SSTD}
Traditional unsupervised SSTD methods are broadly categorized into single-frame and multi-frame approaches. Typical single-frame-based SSTD methods include Hough 
\cite{diprima2018efficient,jiang2022automatic,jiang2022space,cegarra2022real} and Radon transform \cite{hickson2018fast,nir2018optimal} methods, as well as customized stripe-like pattern filtering methods \cite{levesque2007image,levesque2009automatic,dawson2016blind,sara2017faint,virtanen2016streak}. Hough-based and Radon-based methods utilize traditional median filtering \cite{huang1979fast} or morphological filtering techniques \cite{serra1992overview} to eliminate noise and stars, and then employ transform space parameterization to locate stripe-like features. However, these methods are extremely sensitive to noise, fail in low SNR scenarios, and exhibit high computational complexity. Stripe-like pattern filtering methods design stripe filter templates to match stripe-like targets in the images, but they are limited by the variability of stripe-like targets and heavily dependent on parameter settings. Finally, representative multi-frame methods primarily include time index sequence-based approaches \cite{liu2020space1,xi2016space,duarte2023space}. These methods utilize multi-frame information to enhance stripe-like target features while filtering out star points. They then apply multi-level hypothesis testing strategy to identify stripe-like targets among the remaining candidates. However, They are only applicable to sequential space images with a high SNR. Recently, Lin et al. \cite{lin2021new} proposed a method based on stripe-like pattern clustering, but its performance was only reported in slightly noisy scenarios. In summary, traditional unsupervised methods are limited to specific scenarios and exhibit poor generalization in scenarios with variable stripe-like targets, and those with low SNRs affected by noise such as stray light.

\subsection{FSL Methods for Target Detection}
With the rise of CNNs, their application to space images has become a major trend. Jia et al. \cite{jia2020detection} proposed a framework based on Faster R-CNN for classifying point-like stars and stripe-like space targets, marking the first application of CNN in space scenarios. Li et al.\cite{li2022bsc} and Liu et al.\cite{liu2023multi} also developed proprietary CNNs for space stray light removal, demonstrating good performance and further confirming the feasibility of CNNs in space scenarios. Additionally, many FSL networks have been specifically designed to handle low SNRs scenarios affected by complex background noise, including UIU-Net \cite{UIU-Net}, DNANet \cite{DNANET}, ACM \cite{ACM}, RDIAN \cite{RDIAN}, ISNet \cite{ISNET}, APGCNet \cite{AGPCNET}. The above methods all provide new insights into SSTD task. However, their performance relies heavily on complete pixel-level image annotations, which are labor-intensive and challenging to obtain accurately in space images. 

\subsection{SSL Methods for Target Detection}
In recent years, to reduce the dependence on extensive data annotation, numerous SSL approachs have been proposed across various fields, such as medical segmentation \cite{lu2023mutually,wu2023compete}, remote sensing \cite{chen2023semiroadexnet,han2024c2f}, and crack detection \cite{wang2021semi,jian2024cross}. Compared to the FSL methods, these approaches use a large amount of unlabeled data and a small amount of labeled data to enhance the model's generalization capabilities. Currently, commonly used SSL methods fall into two paradigms: consistency regularization-based methods \cite{tarvainen2017mean,sajjadi2016regularization,laine2016temporal,french2019semi} and pseudo-labeling-based methods \cite{liu2021unbiased,lee2013pseudo,sohn2020simple,xu2021end,zhou2023semi}. Consistency regularization, first proposed by Sajjadi et al. \cite{sajjadi2016regularization}, ensures that unlabeled data produce consistent outputs under different data augmentations. This approach has been successfully applied to MT \cite{tarvainen2017mean}, temporal ensembling \cite{laine2016temporal}, and the cutmix-seg architecture \cite{french2019semi}, achieving excellent performance. However, these methods assume consistent model outputs for different perturbed inputs, potentially leading to overconfidence in noisy labels and reducing the ability to accurately learn and generalize from the true data structure.

On the other hand, pseudo-labeling methods use labeled data to pre-train the network, then generate labels for unlabeled data through the model itself, and finally incorporate these pseudo-labels into the training process. This approach includes two iterative aspects: 1) the model generates pseudo-labels, and 2) these pseudo-labels are used to update the model weights, leading to the generation of higher-quality pseudo-labels. The advantage of this approach is its ability to dynamically expand the training dataset, but it relies heavily on the quality of the pseudo-labels. Lee et al. \cite{lee2013pseudo} first proposed this type of method, and then Sohn et al. \cite{sohn2020simple} added strong image augmentation to enhance the model's robustness. To address bias and manual intervention in the pseudo-label generation process, Liu et al. \cite{liu2021unbiased} and Xu et al. \cite{xu2021end} introduced end-to-end strategies based on the MT framework \cite{tarvainen2017mean}. Specifically, they applyed weak data augmentation in the teacher model and strong data augmentation in the student model to achieve a more robust training process. Moreover, they utilized the EMA mechanism to update the weights of the teacher model, effectively incorporates the historical performance of the student model, smoothing the output of the teacher model. Notably, the EMA mechanism introduced in the MT framework \cite{tarvainen2017mean} has also been shown in subsequent studies to be crucial for improving SSL performance \cite{yang2021interactive,mao2023semi,zhou2023semi}.

\section{Collaborative Static-Dynamic Teaching Framework for SSTD}
\label{sec:Methodology}

Our proposed framework comprises three main components: First, we introduced a dual-teacher CSDT SSL architecture. Next, we developed an APL strategy that leverages the strong directional features of stripe-like targets to identify the optimal pseudo-labels. Finally, we designed a network configuration known as MSSA-Net specifically tailored for the SSTD task.


\begin{figure*}[ht]
  \centering
  \includegraphics[width=\textwidth]{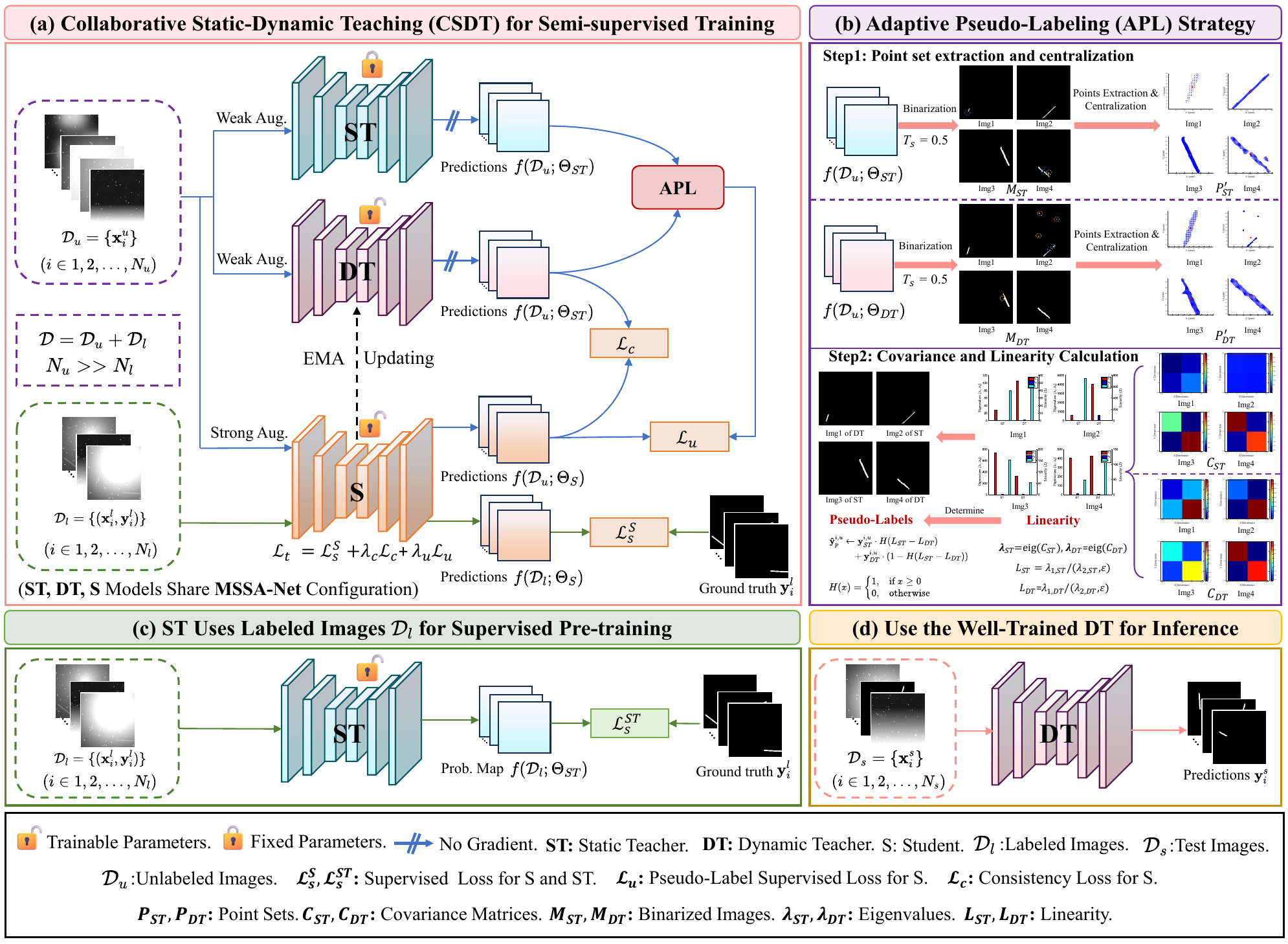}
  \caption{Overview of our proposed CSDT SSL framework. (a) CSDT uses both labeled images $\mathcal{D}_l$ and unlabeled images $\mathcal{D}_u$ for SSL training, where the ST model is a pre-trained teacher whose weights remain fixed during training. The DT and S model are trainable, and the parameters of the DT model are updated through the EMA strategy using the parameters from the S model. (b) Details of the APL strategy implementation, which involves point set extraction and centralization, followed by covariance and linearity calculations. (c) ST uses a small subset of labeled images $\mathcal{D}_l$ for pre-training. (d) During inference, only the well-trained DT model is utilized.}
  \label{fig:fig3}
\end{figure*}

\subsection{CSDT SSL Architecture}
\label{sec:IIIA}
As shown in Fig.~\hyperref[fig:fig3]{\ref*{fig:fig3}(a)}, our proposed CSDT SSL framework includes a ST model, a DT model, and a S model. These models all share a unified network, the MSSA-Net, which is specifically designed for the SSTD task. The ST and DT models guide the S model using the APL strategy. The ST model's weights keep fixed during SSL training, while the S model's weights are updated via gradient backpropagation. The DT model's weights are updated using the EMA mechanism based on the S model. The detailed CSDT SSL training and updating strategies for these models are outlined in Algorithm \ref{alg:CSDT} and described below.

\subsubsection{Pre-Training of the ST Model}It is initially pre-trained on a limited set of labeled space images $\mathcal{D}_l = \{(\mathbf{x}_i^l,\mathbf{y}_i^l) | i \in {1,2, \ldots, N_l}, \mathbf{x}_i^l \subseteq \mathbb{R}^{H \times W \times C}, \mathbf{y}_i^l \subseteq \mathbb{R}^{H \times W}\}$. Here, $\mathbf{x}_i^l$ and $\mathbf{y}_i^l$ represent labeled images and their ground truth (GT), respectively; the parameters $H$, $W$, and $C$ denote the height, width, and number of channels of the images, respectively; while $N_l$ indicates the number of labeled samples. 

To achieve this, samples from $\mathcal{D}_l$ are iteratively fed into the ST model for supervised pre-training using the Dice loss function $l_{d}$, as shown in Fig.~\hyperref[fig:fig3]{\ref*{fig:fig3}(c)}. The supervised loss for the ST model, denoted as $\mathcal{L}_{s}^{ST}$, is calculated as follows:

\begin{equation}
   \mathcal{L}_{s}^{ST}  = \frac{1}{|B_l|} \sum_{(\mathbf{x}_i^l, \mathbf{y}_i^l) \in B_l} l_{d}(f(\mathbf{x}_i^l;\Theta_{ST}), \mathbf{y}_i^l),
\end{equation}
where $B_l$ represents a batch of labeled images fed into the ST model, and $\Theta_{ST}$ are the weight parameters of the ST model.

As shown in Fig.~\hyperref[fig:fig3]{\ref*{fig:fig3}(a)}, during the training phase of the CSDT SSL framework, its weight parameters are fixed, and only unlabeled images $\mathcal{D}_u = \{\mathbf{x}_i^u | i \in {1,2,\ldots,N_u}, \mathbf{x}_i^u \subseteq \mathbb{R}^{H \times W \times C}\}$ are fed into the ST model. Serving as a knowledgeable teacher in SSTD, the ST model plays a primary teaching role in the early stages of the SSL training process, especially as the DT model lacks SSTD capabilities. The main goal of the ST model at this stage is to generate pseudo-labels $\mathbf{y}_{ST}^{i,u} = f(\mathcal{D}_u; \Theta_{ST})$ for the unlabeled images $\mathcal{D}_u$, providing initial guidance to the S model. This role ensures that the S model starts on the correct learning trajectory, effectively setting a performance baseline within the CSDT framework.

\subsubsection{Updating of the DT Model} 
Unlike the ST model whose weights are fixed, the DT model has trainable weights. 
To update its weights, we also employ the EMA mechanism, which is commonly used in the MT framework. This mechanism ensures that the DT model's weights are dynamically and smoothly updated using the historical weight parameters of the S model. The weights update formula for the DT model is as follows:

\begin{equation}
   \Theta_{DT}^t = \alpha\Theta_{DT}^{t-1} + (1-\alpha)\Theta_{S}^{t},
\end{equation}
where $\alpha$ is the decay rate, typically set between 0.9 and 0.999, to control the influence of the student's historical information on the DT model's weights. Following the settings of most papers, the value is set to 0.99 in this study. $\Theta_{DT}$ and $\Theta_{S}$ represent the weight parameters of the DT and S models, respectively, and $t$ denotes the current iteration number.

During the CSDT SSL training, the DT model also only uses unlabeled images $\mathcal{D}_u$ from the dataset, generating pseudo-labels denoted as $\mathbf{y}_{DT}^{i,u} = f(\mathcal{D}_u; \Theta_{DT})$. During inference, only the DT model is used to perform inference on the test dataset $\mathcal{D}_s = \{\mathbf{x}_i^s | i \in {1,2,\ldots,N_s}\}$, as illustrated in Fig.~\hyperref[fig:fig3]{\ref*{fig:fig3}(d)}.


\subsubsection{SSL of the S Model}  Each batch of images fed into the S model consists of both labeled images $\mathcal{D}_l$ and unlabeled images $\mathcal{D}_u$. Due to the significantly higher number of unlabeled images ($N_u$) compared to labeled ones ($N_l$), labeled images are repeatedly sampled during the CSDT SSL training process. As training progresses, the teaching mode shifted from the initial fixed guidance of the ST model to the adaptive collaborative teaching. This adaptive process involves filtering the optimal pseudo-labels for unlabeled space images from both the DT and ST models, thereby enhancing the overall performance of the S model. 

The S model is trained using a customized joint loss function, with weights updated via gradient backpropagation. Specifically, the total loss $\mathcal{L}_{t}$ of the S model comprises three components: the supervised loss $\mathcal{L}_{s}^{S}$, the pseudo-label supervised loss $\mathcal{L}_{u}$, and the consistency loss $\mathcal{L}_{c}$. The supervised loss $\mathcal{L}_{s}^{S}$ is generated using the labeled images $\mathcal{D}_l $ and is calculated as follows:

\begin{equation}
   \mathcal{L}_{s}^S  = \frac{1}{|B_l|} \sum_{(\mathbf{x}_i^l, \mathbf{y}_i^l) \in B_l} l_{d}(f(\mathbf{x}_i^l;\Theta_{S}), \mathbf{y}_i^l),
\end{equation}
where \(l_{d}\) is the Dice loss; \(B_l\) is a batch of labeled images fed into the S model;  and \(\Theta_{S}\) denotes the weight parameters of the S model.

\begin{algorithm}
\caption{CSDT SSL training and updating strategies}
\label{alg:CSDT}
\begin{spacing}{1}
\begin{algorithmic}[1]
\Require ST model $\Theta_{ST}$, DT model $\Theta_{DT}$, S model $\Theta_{S}$
\Require Labeled images $ \mathcal{D}_l = \{{(\mathbf{x}_i^l,\mathbf{y}_i^l)}| i \in \{1,2,...,N_l\}$, Unlabeled images $ \mathcal{D}_u = \{{\mathbf{x}_i^u}| i \in \{1,2,...,N_u\}$
\Require Loss functions $\mathcal{L}_{s}^{ST}, \mathcal{L}_{s}^{S}, \mathcal{L}_{c}, \mathcal{L}_{u}$ 
\Require Loss weight factors $\lambda_c$, $\lambda_u$, EMA decay rate $\alpha$
\Require Adaptive pseudo-labeling strategy $APL(\cdot)$
\Ensure Well-trained $\Theta_{DT}^{(num\_epochs)}$ for SSTD
\State \textit{\textbf{Train:}}
\State  $\Theta_{ST}, \Theta_{DT}, \Theta_{S} \gets \text{init}( )$ \Comment{Initialize all models}
\State $t \gets 0$ 
\Comment{Initialize iteration counter}
\State $\Theta_{ST} \gets \text{train}(\mathbf{x}_i^l,\mathbf{y}_i^l)$ 
\Comment{Pre-train ST with $\mathcal{D}_l$}
\For{epoch = 1 to num\_epochs}
    \For{each minibatch $B_l$ in $\mathcal{D}_l$, $B_u$ in $\mathcal{D}_u$}
        \State $t \gets t + 1$ 
        \For{$\mathbf{x}_i^u \subseteq B_u$}
            \State $\mathbf{y}^{i, u}_{ST} \gets f(\mathbf{x}_i^u; \Theta_{ST})$
            \State $\mathbf{y}^{i, u}_{DT} \gets f(\mathbf{x}_i^u; \Theta_{DT})$
            \State $\mathbf{\hat{y}}_{p}^{i,u} \gets APL(\mathbf{y}^{i, u}_{ST}, \mathbf{y}^{i, u}_{DT})$
            \State $\mathcal{L}_{c} \gets \frac{1}{|B_u|} \sum l_{m}(f(\mathbf{x}_i^u;\Theta_{S}), f(\mathbf{x}_i^u;\Theta_{DT}))$
            \State $\mathcal{L}_{u} \gets \frac{1}{|B_u|} \sum l_{d}(f(\mathbf{x}_i^u;\Theta_{S}), \mathbf{\hat{y}}_{p}^{i,u})$
        \EndFor
        \For{$\mathbf{x}_i^l \subseteq B_l$}
            \State $\mathcal{L}_{s}^S \gets \frac{1}{|B_l|} \sum l_{d}(f(\mathbf{x}_i^l; \Theta_{S}), \mathbf{y}_i^l)$ 
        \EndFor
        \State $\mathcal{L}_{t} \gets \mathcal{L}_{s}^{S} + \lambda_c \mathcal{L}_{c} + \lambda_u \mathcal{L}_{u}$
        \Comment Update $\Theta_{S}^{t}$
        \State $\lambda_c = \exp \left( -5 \left(1 - t / t_{\max} \right)^2 \right)$
        \Comment Ramp-up weight
        \State $\Theta_{DT}^{t} \gets \alpha \Theta_{DT}^{t-1} + (1 - \alpha) \Theta_{S}^{t}$ \Comment{EMA Updating }
    \EndFor
\EndFor
\State \textit{\textbf{Inference:}}
\For{each $\mathbf{x}_i^s \subseteq \mathcal{D}_s$}
    \State $\mathbf{y}_i^s \gets f(\mathbf{x}_i^s; \Theta_{DT}^{(num\_epochs)})$ \Comment{Predict with $\Theta_{DT}$}
\EndFor
\end{algorithmic}
\end{spacing}
\end{algorithm}

The consistency loss $\mathcal{L}_{c}$ ensures that the S model's prediction aligns with the DT model's under different augmentation conditions, maintaining uniformity in predictions for unlabeled space images $\mathcal{D}_u $. In this paper, we apply colour jittering and Gaussian blur, both at varying intensities, for strong and weak image augmentation, respectively. The consistency loss is calculated as follows:

\begin{equation}
   \mathcal{L}_{c}  = \frac{1}{|B_u|} \sum_{\mathbf{x}_i^u \in B_u} l_{m}(f(\mathbf{x}_i^u;\Theta_{S}), f(\mathbf{x}_i^u;\Theta_{DT})),
\end{equation}
where \(B_u\) is a batch of unlabeled images fed into the S model; \(\mathbf{x}_i^u\) denotes the unlabeled space images; \(l_{m}\) represents the mean squared error (MSE) loss.

The pseudo-label supervised loss $\mathcal{L}_{u}$ leverages high-quality pseudo-labels provided by either the ST model or the DT model for learning from unlabeled space images. This loss is specifically designed to uncover deeper insights into the spatial patterns and characteristics of stripe-like targets within the unlabeled space images, thus improving the generalization performance of the S model. It is computed as follows:
 
\begin{equation}
   \mathcal{L}_{u}  = \frac{1}{|B_u|} \sum_{\mathbf{x}_i^u \in B_u} l_{d}(f(\mathbf{x}_i^u;\Theta_{S}), \mathbf{\hat{y}}_{p}^{i,u}),
\end{equation}
where \(l_{d}\) is the Dice loss, and \(\mathbf{\hat{y}}_{p}^{i,u}\) is the optimal pseudo-label generated by either the ST model or DT model, expressed as $\mathbf{\hat{y}}_{p}^{i,u} = APL(\mathbf{y}_{ST}^{i,u},\mathbf{y}_{DT}^{i,u})$. 

The joint loss \(\mathcal{L}_{t}\) for the S model is the weighted sum of these individual losses, expressed as follows:

\begin{equation}
   \mathcal{L}_{t} = \mathcal{L}_{s}+ \lambda_c\mathcal{L}_{c} + \lambda_u\mathcal{L}_{u},
\end{equation}
where \(\lambda_u\) is a fixed constant, and \(\lambda_c\) is the ramp-up weight, dynamically adjusting the contribution of the consistency loss \(\mathcal{L}_{c}\) to the total loss during training, which is defined as:
\begin{equation}
    \lambda_c = \exp \left( -5 \left(1 - t / t_{\max} \right)^2 \right),
\end{equation}
where \(\exp(\cdot)\) denotes the exponential function; \(t\) represents the current iteration number; and \(t_{\max}\) is the maximum number of iterations.

Initially, \(\mathcal{L}_{s}\) and \(\mathcal{L}_{u}\) dominate the loss composition. As training progresses, the weight of \(\mathcal{L}_{c}\) gradually increases, reflecting its growing importance in fine-tuning the model’s performance. The combination of these losses enables the S model to effectively learn from both labeled and unlabeled space images, thereby enhancing its ability to accurately detect stripe-like targets. Moreover, the performance improvement of the S model further amplifies the SSTD capability of the DT through the EMA mechanism.

\subsection{APL Strategy}
\label{sec:IIIB}

As mentioned earlier, at the beginning of the SSL training, the pseudo-labels generated by the ST model are more reliable due to its pre-training. However, the DT model's ability to segment stripe-like targets is quite weak at this stage, resulting in noisy prediction maps. As training progresses, the DT model gradually improves its performance by leveraging insights into the distribution of unlabeled space images. Therefore, we initially used the number of connected components as a straightforward metric to assess whether the DT model possesses any stripe-like target segmentation capability. 

As illustrated in Fig.~\hyperref[fig:fig3]{\ref*{fig:fig3}(b)}, we first apply a binarization threshold $T_s = 0.5$ to the Sigmoid predictions of the ST and DT models to filter out low-confidence predictions, converting them into binary maps $M_{ST} = f(\mathcal{D}_u;\Theta_{ST})>T_s$ and $M_{DT} = f(\mathcal{D}_u;\Theta_{DT})>T_s$. We then calculate the numbers of connected components $N_{ST}$ and $N_{DT}$ in $M_{ST}$ and $M_{DT}$, respectively. Next, we define the threshold $T_c=\{\max(N_{ST}^i)| i \in {1,2, \ldots, N_u}\}$, based on the maximum number of connected components in all $M_{ST}$. If $N_{DT} > T_c$, the DT model is considered to lack stripe segmentation capability, and the pseudo-labels $\mathbf{y}_{ST}^{i,u}$ generated by the ST model are directly used to supervise the S model. Conversely, if $N_{DT} < T_c$, the pseudo-labels are selected based on the predictions with higher-quality stripe-like targets from either the ST or DT models.

To facilitate this selection, we define a new linearity metric for each prediction based on the fact that each stripe-like target aligns along a specific direction. This metric helps adaptively select the optimal pseudo-labels from either the ST or DT models. We refer to this process as adaptive pseudo-labeling (APL) strategy. The overall process of the proposed APL strategy is detailed in Algorithm \ref{alg:APL}, which can be briefly summarized as follows: 

\begin{equation}
    \mathbf{\hat{y}}_{p}^{i,u} =
    \begin{cases} 
        \mathbf{y}_{ST}^{i,u}, & \text{if } N_{DT} > T_c \\
        APL(\mathbf{y}_{ST}^{i,u}, \mathbf{y}_{DT}^{i,u}). & \text{otherwise}
    \end{cases}
\end{equation}

\begin{algorithm}
\caption{APL strategy}
\label{alg:APL}
\begin{algorithmic}[1]
\Require Unlabeled images $ \mathcal{D}_u = \{{\mathbf{x}_i^u}| i \in \{1,2,...,N_u\}$, predictions of ST, $ \mathbf{y}^{i, u}_{ST}$, and DT, $ \mathbf{y}^{i, u}_{DT}$
\Require Binarization and connectivity threshold, $T_s$ and  $T_c$
\Ensure Optimal pseudo-label $\mathbf{\hat{y}}_{p}^{i,u}$
\State $B(x) (x > T_s)$ \Comment{Binarization function}
\State $CC(M)  $ \Comment{Number of connected components in $M$}
\State
\[
H(x) = 
\begin{cases} 
1, & \text{if } x \geq 0 \\
0, & \text{otherwise}
\end{cases}
\]

\Function{calculate\_linearity}{$M$}
    \State $P \gets \{ (x, y) \mid M[x,y] > 0 \}$ 
    \Comment{Get the point set}
    \State $s \gets \frac{1}{|P|} \sum_{p \in P} p$
    \Comment{Get the centroid}
    \State $P' \gets \{ p - s \mid p \in P \}$
    \Comment{Centralization}
    \State $C \gets \frac{1}{|P'|} P'^T P'$
    \Comment{Get the covariance matrix}
    \State $ \boldsymbol{\lambda} \gets \text{eig}(C)$ 
    \Comment{Compute the eigenvalues}
    \State $ \boldsymbol{\lambda}_{sorted} \gets \text{sort}( \boldsymbol{\lambda}, \text{descend})$ \Comment{Sort descending}
    \State $\lambda_{1,*} \gets \boldsymbol{\lambda}_{sorted}[0]$ \Comment{The largest eigenvalue}
    \State $\lambda_{2,*} \gets \boldsymbol{\lambda}_{sorted}[1]$ \Comment{The second largest eigenvalue}
    \State \Return $\frac{\lambda_1}{\max(\lambda_2,\varepsilon)}$
\EndFunction

\For{$\text{epoch} = 1$ \textbf{to} $\text{num\_epochs}$}
    \For{$\mathbf{x}_i^u \subseteq B_u$}
        \State $M_{ST} \gets B(\mathbf{y}_{ST}^{i,u})(\mathbf{y}_{ST}^{i,u}>0)$ 

        \State $M_{DT} \gets B(\mathbf{y}^{i, u}_{DT})(\mathbf{y}^{i, u}_{DT}>0)$ 

        \State $N_{ST} \gets CC(M_{ST})$ 
        \State $N_{DT} \gets CC(M_{DT})$ 
        \If{$|N_{ST} - N_{DT}| > T_c$}
            \State $\mathbf{\hat{y}}_{p}^{i,u} \gets \mathbf{y}^{i, u}_{ST}$
        \Else
            \State $L_{ST} \gets \Call{calculate\_linearity}{M_{ST}}$
            \State $L_{DT} \gets \Call{calculate\_linearity}{M_{DT}}$
            \State $\mathbf{\hat{y}}_{p}^{i,u} \gets \mathbf{y}^{i, u}_{ST} \cdot H(L_{ST} - L_{DT}) + \mathbf{y}^{i, u}_{DT} \cdot (1 - H(L_{ST} - L_{DT}))$

        \EndIf
        \State train$(\mathbf{x}_i^u, \mathbf{\hat{y}}_{p}^{i,u})$
        \Comment{Training with pseudo-label $ \mathbf{\hat{y}}_{p}^{i,u} $}
    \EndFor
\EndFor
\end{algorithmic}
\end{algorithm}

Specifically, the steps of APL strategy involve extracting the pixel areas with non-zero pixel values in the binary images $M_{ST}$ and $M_{DT}$ respectively, and then to form the point sets $P_{ST}$ and $P_{DT}$ as follows:

\begin{equation}
\begin{aligned}
    P_{ST} &= \left\{(x, y) \in \mathbb{R}^2 \mid M_{ST}[x,y] > 0 \right\}, \\
    P_{DT} &= \left\{(x, y) \in \mathbb{R}^2 \mid M_{DT}[x,y] > 0 \right\}.
\end{aligned}
\end{equation}

These two point sets represent the stripe-like target areas predicted by the ST and DT models, respectively. First, we calculate the centroid coordinates for each point set using the equations \(s_{ST} = \frac{1}{|P_{ST}|} \sum_{p \in P_{ST}} p\) and \(s_{DT} = \frac{1}{|P_{DT}|} \sum_{p \in P_{DT}} p\), which are denoted by the red symbols '\textcolor{red}{$\times$}' in Fig.~\hyperref[fig:fig3]{\ref*{fig:fig3}(b)}. We then perform centralisation preprocessing on each point set to reduce the external differences caused by position deviation and to focus on the internal shape distribution of the stripe-like target. The newly obtained point sets, \(P'_{ST}\) and \(P'_{DT}\), are expressed as follows:

\begin{equation}
    \begin{aligned}
        P'_{ST} &= \{p - s_{ST} \mid p \in P_{ST}\}, \\
        P'_{DT} &= \{p - s_{DT} \mid p \in P_{DT}\}.
    \end{aligned}
\end{equation}

Then, we construct the covariance matrices for the new point sets \( P'_{ST} \) and \( P'_{DT} \) as: $C_{ST} = \frac{1}{|P'_{ST}|} (P'_{ST})^T P'_{ST}, C_{DT} = \frac{1}{|P'_{DT}|} (P'_{DT})^T P'_{DT}$, respectively. The eigenvalues of these covariance matrices are computed as follows:

\begin{equation}
    \begin{aligned}
    \boldsymbol{\lambda}_{ST} &= \text{eig}(C_{ST}),\\ 
    \boldsymbol{\lambda}_{DT} &= \text{eig}(C_{DT}), \\
    \end{aligned}
\end{equation}
where eig($\cdot$) means to obtain the eigenvalue of $P'_{ST}$ and $P'_{DT}$. The eigenvalues are arranged in descending order and recorded as $\boldsymbol{\lambda}_{ST} = \{\lambda_{1,ST}, \lambda_{2,ST}\}$ and $\boldsymbol{\lambda}_{DT} = \{\lambda_{1,DT}, \lambda_{2,DT}\}$. 

Since stripe-like targets exhibit strong directional consistency, a larger $\lambda_{1,*}$ indicates that the model more effectively captures the primary extension direction of stripe-like targets in predicted images. Consequently, diffusion in the width direction should be minimal, necessitating that $\lambda_{2,*}$ be kept as small as possible. Based on these eigenvalues, we define the linearity \(L_*\) of each predicted image as follows:

\begin{equation}
    \begin{aligned}
    L_{ST} = \frac{\lambda_{1,ST}}{\max(\lambda_{2,ST}, \varepsilon)},\\
    L_{DT} = \frac{\lambda_{1,DT}}{\max(\lambda_{2,DT}, \varepsilon)},\\
    \end{aligned}
\end{equation}
where $\varepsilon$ is an infinite decimal to prevent the denominator from being zero, and $\max$($\cdot$) represents the maximum function. 

From the analysis above, a higher \(L_*\) value indicates better directionality within the point set, suggesting that the detected stripe-like targets are of higher quality and more appropriate for use as pseudo-labels. 
Consequently, during training, the CSDT SSL framework adaptively identifies the optimal pseudo-labels for unlabeled space images by comparing the linearity metrics \(L_*\) of the predictions from the ST and DT models as follows:

\begin{equation}
\begin{aligned}
    \mathbf{\hat{y}}_{p}^{i,u} &= APL(\mathbf{{y}}_{ST}^{i,u}, \mathbf{{y}}_{DT}^{i,u}) \\
    &= \mathbf{y}^{i, u}_{ST} \cdot H(L_{ST} - L_{DT}) + \\
    & \mathbf{y}^{i, u}_{DT} \cdot (1 - H(L_{ST} - L_{DT})),
\end{aligned}
\end{equation}
where $H(x)$ is a step function defined as follows:

\begin{equation}
H(x) = 
\begin{cases} 
1, & \text{if } x \geq 0 \\
0. & \text{otherwise}
\end{cases}
\end{equation}

In Fig.~\hyperref[fig:fig3]{\ref*{fig:fig3}(b)}, we display four unlabeled space images (img1 to img4) from a batch to exemplify the proposed APL strategy. This visualisation clearly shows how the APL strategy adaptively selects the optimal quality predictions from the ST or DT models to serve as the pseudo-labels, denoted $\mathbf{\hat{y}}_{p}^{i,u}$.

\subsection{MSSA-Net Configuration}
\label{sec:IIIC}
As the segmentation network used by the ST, DT, and S models in CSDT framwork, the MSSA-Net is illustrated in Fig.~\ref{fig:fig4}, consisting of two primary stages: multi-scale  feature extraction endoer and multi-level feature fusion decoder. During feature extraction endoer stage, we employ a MDPC block and a FMWA block, explicitly tailored to detect stripe-like patterns. These designs enhance the network's sensitivity to various scales of the stripe-like targets, improving its ability to discern both local and global details critical for SSTD task. In the decoder stage, multi-level features are fused to improve feature representation, effectively detecting stripe-like space targets accurately. 



\subsubsection{Multi-Scale Feature Extraction Stage}
\begin{figure*}[ht]
  \centering
  \includegraphics[width=\textwidth]{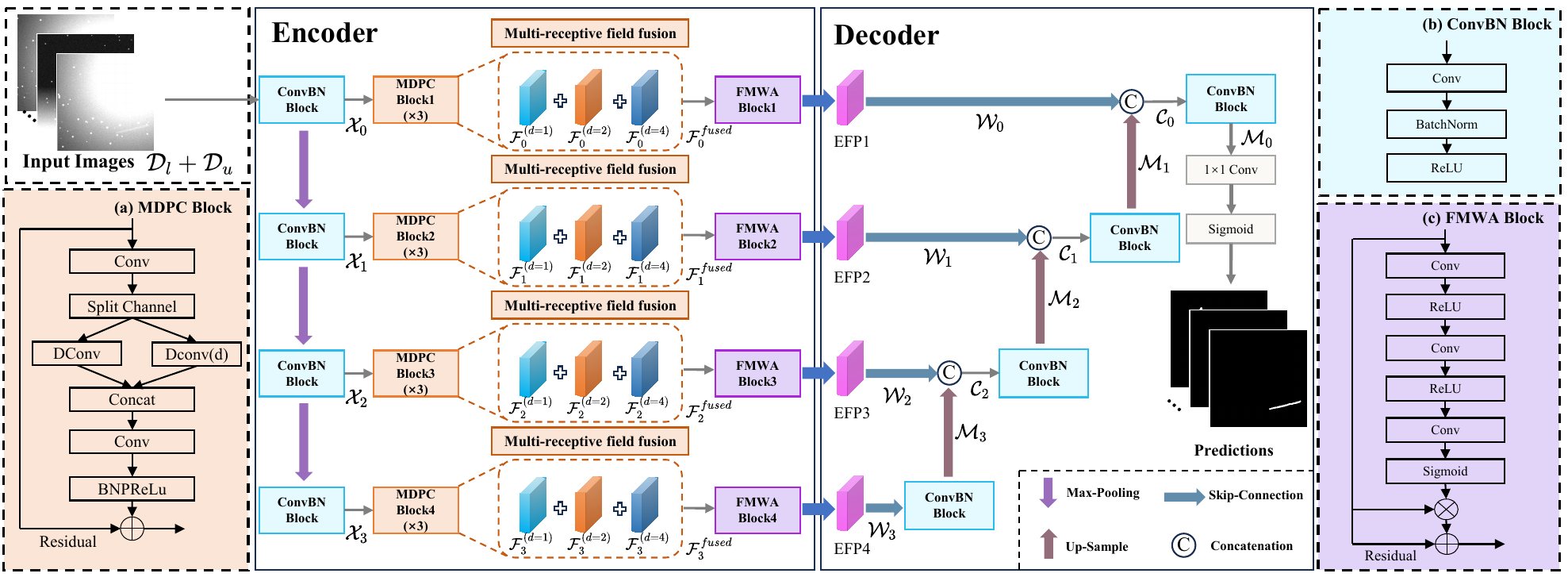}
  \caption{Overall configuration of MSSA-Net comprises two main parts: the encoder and the decoder. The encoder is used for multi-scale stripe feature extraction, while the decoder engages in multi-level feature fusion. (a) The internal architecture of the proposed MDPC block, designed to expand the receptive field and extract multi-scale stripe features. (b) The internal architecture of the ConvBN block. (c) The internal architecture of the proposed FMWA block, which enhances the stripe-like target regions in the feature map through feature reconstruction and suppresses noise.}
  \label{fig:fig4}
\end{figure*}

Stripe-like targets exhibit marked variability and low SNR in various space images, which often leads to breakage within target regions detected by existing CNN methods. In some cases, these targets may not be detected at all because of the methods' limited capacity to adapt to the diverse feature scales and faint textures of stripe-like targets. To address this issue, we incorporate multiple MDPC blocks after each downsampling step in the feature extraction phase. Each MDPC block is designed with different dilation rates to enhance the network's adaptability to space resolutions at various scales. This design enables MSSA-Net to effectively integrate a broad spectrum of stripe feature information, ensuring that the final feature maps capture both intricate local details and extensive contextual information. As shown in Fig.~\hyperref[fig:fig3]{\ref*{fig:fig4}(a)}, each MDPC block contains two different convolution paths, enhancing the flexibility and efficiency of feature extraction. Initially, $\mathcal{X}_i$ undergoes a standard $3\times3$ convolution, producing a feature map while reducing the number of input channels to $C/2$. Subsequently, the feature map is split into two branches, $\mathcal{F}_{left}$ and $\mathcal{F}_{right}$, as follows:

\begin{equation}
    \begin{aligned}
        \mathcal{F}_{left} &= \text{DConv}(\text{Split}(\text{Conv}(\mathcal{X}_i))), \\
        \mathcal{F}_{right} &= \text{DConv}_{d}(\text{Split}(\text{Conv}(\mathcal{X}_i)),d),
    \end{aligned}
\end{equation}
where $\mathcal{X}_i$ represents the feature output after the $i$-th downsampling; $\text{Conv}(\cdot)$ denotes the standard $3\times3$ convolution; $\text{DConv}(\cdot)$ refers the depthwise convolution; $\text{DConv}_d(\cdot)$ indicates the depthwise dilated convolution with the specified dilation rate $d$; and $\text{Split}(\cdot)$ describes the channel splitting of the feature map. 

At this stage, the output channels of the left and right branches are $C/4$ each. Specifically, the left path uses depth-wise convolution to precisely capture details of the short stripe-like targets, while the right path employs depth-wise dilated convolution to expand the receptive field and capture broader contextual information. This dual-path design not only enhances the network's ability to discern the texture information of stripe-like space targets but also improves its capacity to integrate these details over a larger spatial range, thereby significantly enhancing the overall accuracy of the SSTD task. After processing through these two paths, the features are fused and further refined to enhance the stripe saliency of the feature map:

\begin{equation}
    \begin{aligned}
    \mathcal{F}_{concat} = \text{Conv}(\mathcal{F}_{left} \oplus \mathcal{F}_{right}), \\
    \end{aligned}
\end{equation}
where $\oplus$ represents Concatenation. At this stage, the number of channels of the concatenated feature map ($\mathcal{F}_{left} \oplus \mathcal{F}_{right}$) is restored to $C/2$. Then, a $3\times3$ convolution restores the number of channels to the original $C$, while $\mathcal{F}_{concate}$ extracts more extensive contextual information of the space images than the original input feature $\mathcal{X}_i$. 

Next, $\mathcal{F}_{concat}$ is batch normalised (BN) and activated via the PReLU function. It is then fused with the original input feature $\mathcal{X}_i$ through a residual connection to produce the output $F_i^d$ of each MPDC block. These steps ensure continuity in feature representation and enhance the network's gradient propagation:

\begin{equation}
    \mathcal{F}_i^d = \text{BNPReLU}(\mathcal{F}_{concat} + \mathcal{X}_i), \quad \forall d \in \{1, 2, 4\}.
\end{equation}

Then, at each downsampling step, the multi-scale features extracted by each MDPC block under different receptive fields are fused to produce the composite feature map $\mathcal{F}_i^{fused}$:

\begin{equation}
    F_i^{fused} = \sum_{d \in \{1, 2, 4\}} F_i^d, \quad \forall i  \in \{0, 1, 2, 3\}.
\end{equation}

Finally, in each downsampling step, we also incorporate a FMWA block to further enhance the response of the stripe-like pattern area in $\mathcal{F}_i^{fused}$ and suppress irrelevant features. Its architecture is shown in Fig.~\hyperref[fig:fig4]{\ref*{fig:fig4}(c)}, which performs feature transformation through a series of convolution and ReLU layers to enhance the features in a nonlinear manner. A Sigmoid activation layer then generates a weighted attention map, ranging from 0 to 1, to signify the importance of each feature. This weighted attention map is then applied to the input feature map $\mathcal{F}_i^{fused}$, dynamically reconstructing important features and amplifying their impact on the final model prediction. Additionally, this block includes residual connections that add the original input $\mathcal{F}_i^{fused}$ to the weighted output, preserving the integrity of the stripe-like pattern and improving the stability of training. The resulting each enhanced feature map (EFP), denoted as $\mathcal{W}_i$, is calculated as follows:

\begin{equation}
    \mathcal{W}_i = \sigma(\text{Conv}_{deep}(\mathcal{F}_i^{fused})) \odot \mathcal{F}_i^{fused} + \mathcal{F}_i^{fused},
\end{equation}
where \(\text{Conv}_{deep}\) represents a deep feature extractor composed of multiple convolutional and ReLU layers; $\sigma(\cdot)$ denotes a Sigmoid function; and $\odot$ signifies pixel-level multiplication.

\subsubsection{Multi-level Feature Fusion Stage}
We employ ConvBN blocks as shown in Fig.~\hyperref[fig:fig4]{\ref*{fig:fig4}(b)} to refine feature maps at each depth of MSSA-Net. Specifically, the enhanced feature maps $\mathcal{W}_i$ from the FMWA blocks in each downsampling stage are sequentially fed into each decoder layer. We then merge up-sampled features $\mathcal{M}_{i+1}$ with the skip-connected enhanced features $\mathcal{W}_i$ to reintroduce essential spatial details from early high-resolution layers, producing comprehensive combined feature maps $\mathcal{C}_i$. This integration of low-level stripe-like texture information with high-level semantic insights is crucial for restoring textures lost during downsampling. The resulting fused image $\mathcal{M}_0$, which matches the size of the original image, is subsequently processed by a 1x1 convolution and a Sigmoid activation function $\sigma(\cdot)$ to produce accurate predictions of the stripe-like target locations. The approach of this multi-level feature fusion module ensures robust detection capabilities for stripe-like space targets across various scales and directions, thus enhancing the reliability in practical applications.

\section{Experiments}
\label{sec:exper}
This section presents a thorough evaluation of the proposed CSDT SSL framework. It starts by outlining the experimental setup, including the datasets, evaluation metrics, and implementation details. It then provides a comprehensive analysis through quantitative and visual comparisons with SOTA SSL methods across various network configurations. Additionally, the section includes extensive ablation studies to assess the CSDT framework’s generalization capabilities and effectiveness of its key components, such as the zero-shot generalization capabilities of the well-trained SSL models on diverse real-world space image datasets; the effectiveness of MSSA-Net components; the roles of different teacher models in the CSDT; evaluation of the APL strategy; and the impact of various loss function combinations. These experiments collectively underscore the contributions of individual elements to the overall performance of the proposed framework.

\subsection{Experimental Setup}
This section outlines the experimental setup, starting with an overview of the datasets used in this study, including its composition and the noise challenges it presents. It then discusses the evaluation metrics used to measure model performance at both the pixel and target levels. Finally, it details the implementation, covering hardware and software configurations, training parameters, and specific loss functions, to ensure clarity and reproducibility.

\subsubsection{Dataset} We conducted extensive experiments on the AstroStripeSet \cite{zhu2024sstd}, which contains stripe-like space target images challenged by four types of space stray light noise: earthlight, sunlight, moonlight, and mixedlight. It includes 1,000 fixed training images, 100 fixed validation images, and 400 fixed test images, with each stray light noise contributing 100 images to the test set. Notably, nearly all previously published papers have only validated their performance on their in-house datasets, neglecting to use data from other studies to verify the generalization capabilities. To this end, our test dataset incorporated real-world images containing stripe-like space targets that have been used in previous studies \cite{lin2021new}, \cite{li2022bsc}, \cite{liu2023multi}. Since their original images were not publicly available, we used screenshots from published papers to demonstrate our framework's effectiveness on these datasets. We recognized that screenshots could degrade image quality, but this was the only viable way to obtain these images for our study. To enhance image quality, we applied sharpening filters and contrast adjustments. The dataset also included real-world images with stripe-like space targets collected from the Internet, as well as real-world background images captured by our on-orbit space-based cameras and ground-based telescopes. These diverse sources were used to assess the model's zero-shot generalization capabilities, testing its effectiveness across a variety of unseen space scenarios.

\subsubsection{Evaluation Metrics} We used the Dice coefficient \cite{wang2022uctransnet} and mIoU \cite{ma2023msma} metrics to quantitatively evaluate the pixel-level detection performance of the models. These metrics are defined as follows:
\begin{equation}
    \begin{aligned}
        \text{mIoU} & = \frac{1}{N} \sum_{i=1}^{N} \frac{\mathbf{y}_i^s \cap \mathbf{y}_i^{gt}}{\mathbf{y}_i^s \cup \mathbf{y}_i^{gt}}, \\
        \text{Dice} & = \frac{1}{N} \sum_{i=1}^{N} \frac{2 \times |\mathbf{y}_i^s \cap \mathbf{y}_i^{gt}|}{|\mathbf{y}_i^s| + |\mathbf{y}_i^{gt}|},
    \end{aligned}
\end{equation}
where $\mathbf{y}_i^s$ and $\mathbf{y}_i^{gt}$ represent the model predictions and GT of the test images in $\mathcal{D}_s$, respectively.

Furthermore, we used widely recognized target detection metrics, including detection rate (\(P_d\)) and false alarm rate (\(F_a\)), to evaluate the target-level performance of the models \cite{DNANET,ACM,AGPCNET}. \(P_d\) measured the probability of correctly detecting targets within a test image subset, where detection was considered successful only if the IoU between the prediction $\mathbf{y}_i^s$ and GT $\mathbf{y}_i^{gt}$ exceeded 0.5. Meanwhile, \(F_a\) quantified the proportion of false detections, defined as follows:

\begin{equation}
    \begin{aligned}
        P_d = N_d/N_t, \\
        F_a = N_f/N_p,
    \end{aligned}
\end{equation}
where $N_d$ is the number of correctly detected targets, and $N_t$ is the total number of real targets; $N_f$ represents the number of falsely detected pixels, while $N_p$ denotes the total number of pixels in the test image. 

\subsubsection{Implementation Details} 
All experiments were conducted on a RTX 4080 GPU with 16GB memory, using PyTorch version 2.2.0 and CUDA version 12.1. During training, we set the batch size $B_l$ and $B_u$ to 8 for both labeled and unlabeled images, the maximum number of iterations to 25,000, the learning rate to \(1 \times 10^{-4}\), and used the Adam optimizer for weight updates. The segmentation losses \(\mathcal{L}_s\) and \(\mathcal{L}_u\) utilize the Dice loss, while the consistency loss \(\mathcal{L}_c\) employs the MSE loss. Additionally, the loss weight \(\lambda_u\) is fixed at 0.3, and \(\lambda_c\) is a ramp-up weight. During inference, only the DT model is used for prediction. 

\subsection{Comparison with Existing SOTA SSL Methods} 
In this section, we evaluated the proposed CSDT framework by comparing it with several classic and widely-used SSL methods to highlight its advantages. The SSL methods compared included MT \cite{tarvainen2017mean}, UT \cite{liu2021unbiased}, ISMT \cite{yang2021interactive}, PLMT \cite{mao2023semi}, and ST \cite{yang2022st}. Additionally, we performed comparative experiments with well-known networks such as UCTransNet \cite{wang2022uctransnet},  UNet \cite{ronneberger2015u} and our proposed MSSA-Net within these SSL methods to highlight MSSA-Net’s superior SSTD performance. Table \ref{table:table1} presents a comparison of the parameter counts and inference times for the three networks. The proposed MSSA-Net not only enhances network performance but also maintains competitive inference speeds comparable to UNet, while requiring significantly fewer parameters than UCTransNet. 


\begin{table}[h!]
    \centering
    \small
    \caption{Comparison of parameter count and inference time across three networks: UNet, UCTransNet, and MSSA-Net.}
    \label{table:table1}
    \renewcommand{\arraystretch}{1.4}
    \begin{tabularx}{\linewidth}{ Y | Y | Y }
        \midrule
        Network & Parameters  & Inference Time\\ 
        \midrule
        \midrule
        UNet \cite{ronneberger2015u} & 15.0 M& 20 ms\\
        UCTransNet \cite{wang2022uctransnet} & 66.5 M& 32 ms\\
        MSSA-Net (Ours)  & 32.0 M& 22 ms\\ 
        \midrule
        \midrule
    \end{tabularx}
\end{table}

\begin{table*}[t!]
    \centering
    \caption{Performance comparison of the proposed CSDT architecture with SOTA SSL methods on the AstroStripeSet, across various labeling rates and network configurations. Metrics evaluated include Dice $\uparrow$(\%), mIoU  $\uparrow$(\%), $P_d$ $\uparrow$(\%), and $F_a$ $\downarrow$($×10^{-4}$).}
    \label{table:table2}
    \renewcommand{\arraystretch}{1.4}
    \resizebox{1.0\textwidth}{!}{
    \begin{tabular}{c| c| c| c c c c| c c c c| c c c c} 
        \midrule
        \multirow{2}{*}{\textbf{Network}} & \multirow{2}{*}{\textbf{Method}} & \multirow{2}{*}{\textbf{Source}} & \multicolumn{4}{c|}{\textbf{1/4 (250)}} & \multicolumn{4}{c|}{\textbf{1/8 (125)}} & \multicolumn{4}{c}{\textbf{1/16 (62)}}  \\  \cline{4-15}
        & & & \textbf{Dice} & \textbf{mIoU} & \textbf{P}\textsubscript{\textbf{d}} & \textbf{F}\textsubscript{\textbf{a}} & \textbf{Dice} & \textbf{mIoU} & \textbf{P}\textsubscript{\textbf{d}} & \textbf{F}\textsubscript{\textbf{a}} & \textbf{Dice} & \textbf{mIoU} & \textbf{P}\textsubscript{\textbf{d}} & \textbf{F}\textsubscript{\textbf{a}} \\
        \midrule
        \midrule
        \multirow{7}{*}{\makecell{\textbf{UNet} \\ \cite{ronneberger2015u}}} 
        & Sup.only \cite{ronneberger2015u} & MICCAI (2015) & 77.07 & 68.82 & 79.75 & 3.44 & 72.20 & 62.72 & 71.50 & 3.22 & 58.0 & 48.45 & 55.25 & 2.50  \\
        & MT \cite{tarvainen2017mean} & Neurips (2017) & 81.72 & 73.04 & 84.75 & \textbf{3.19} & 77.34 & 68.25 & 78.50 & 2.96 & 72.27 & 62.74 & 73.0 & 3.18  \\
        & UT \cite{liu2021unbiased} & ICLR (2021) & 81.22 & 73.29 & 88.25 & 6.32 & 79.86 & 70.29 & 84.75 & 7.39 & 74.18 & 63.81 & 75.75 & 11.74  \\
        & ISMT \cite{yang2021interactive} & CVPR (2021) & 83.27 & 74.86 & 88.25 & 3.78 & 79.60 & 71.27 & 86.0 & 5.92 & 73.19 & 62.46 & 73.75 & 16.73 \\
        & PLMT \cite{mao2023semi} & ICASSP (2023) & 79.09 & 70.74 & 83.0 & 4.05 & 72.83 & 63.22 & 73.75 & 3.86 & 64.72 & 54.36 & 61.0 & 3.60  \\
        & ST \cite{yang2022st} & CVPR (2022) & 82.39 & 73.86 & 85.50 & 3.25 & 75.06 & 65.61 & 74.75 & \textbf{2.02} & 63.58 & 53.47 & 60.75 & \textbf{2.22}  \\
        & \textbf{CSDT} & \textbf{Ours} &  \textbf{83.35} & \textbf{75.72} & \textbf{89.75} & 3.67 & \textbf{81.36} & \textbf{72.52} & \textbf{86.50} & 3.45 & \textbf{76.72} & \textbf{66.96} & \textbf{82.0} & 3.37  \\

        \midrule
        \midrule
        \multirow{7}{*}{\makecell{\textbf{UCTransNet} \\ \cite{wang2022uctransnet}}}
        & Sup.only \cite{wang2022uctransnet} & AAAI (2022) & 81.04 & 72.36 & 86.50 & 4.48 & 74.92 & 65.94 & 78.50 & 3.87 & 60.21 & 50.65 & 59.0 & 2.86 \\
        & MT \cite{tarvainen2017mean} & Neurips (2017) & 84.99 & 75.98 & 91.0 & 6.73 & 76.41 & 66.85 & 79.25 & 4.01 & 71.86 & 61.71 & 72.0 & 3.69  \\
        & UT \cite{liu2021unbiased} & ICLR (2021) & 83.99 & 74.68 & 91.75 & 5.88 & 83.02 & 73.41 & 90.50 & 5.49 & 76.43 & 66.14 & 78.25 & 8.21  \\
        & ISMT \cite{yang2021interactive} & CVPR (2021) & 83.65 & 74.63 & 90.0 & 4.84 & 80.67 & 70.99 & 87.0 & 4.77 & 77.16 & 66.72 & 81.75 & 4.03  \\
        & PLMT \cite{mao2023semi} & ICASSP (2023) & 82.20 & 72.78 & 89.0 & 4.63 & 75.90 & 66.85 & 81.0 & 4.0 & 61.68 & 52.04 & 60.25 & 3.79  \\
        & ST \cite{yang2022st} & CVPR (2022) & 82.82 & 74.64 & 89.25 & \textbf{3.81} & 76.95 & 67.44 & 79.0 & \textbf{2.76} & 56.43 & 46.85 & 53.75 & \textbf{1.40}  \\
        & \textbf{CSDT} & \textbf{Ours} & \textbf{85.70} & \textbf{77.09} & \textbf{92.0} & 4.56 & \textbf{83.39} & \textbf{73.92} & \textbf{90.75} & 4.13 & \textbf{78.72} & \textbf{68.36} & \textbf{83.25} & 4.17 \\

        \midrule
        \midrule      
        \multirow{7}{*}{\textbf{\makecell{MSSA-Net \\ (Ours)}}} 
        & Sup.only & \textbf{Ours} & 84.98 & 76.76 & 92.25 & 4.61 & 78.31 & 70.01 & 84.25 & 4.50 & 69.69 & 60.09 & 71.25 & 7.93 \\  
        & MT \cite{tarvainen2017mean} & Neurips (2017)  & 85.35 & 77.54 & 91.25 & 4.95 & 80.20 & 71.38 & 85.0 & 4.22 & 73.73 & 64.37 & 75.25 & \textbf{3.95} \\ 
        & UT \cite{liu2021unbiased} & ICLR (2021)  & 85.42 & 77.50 & 92.50 & 6.19 & 83.73 & 75.37 & \textbf{92.25} & 6.38 & 77.98 & 68.49 & 83.25 & 13.43 \\ 
        & ISMT \cite{yang2021interactive}& CVPR (2021)  & 84.31 & 76.67 & 91.75 & 5.66 & 82.34 & 74.14 & 88.25 & 4.69 & 78.68 & 69.20 & 84.0 & 9.94 \\ 
        & PLMT \cite{mao2023semi} & ICASSP (2023) & 83.76 & 75.94 & 91.75 & 4.94 & 80.57 & 71.65 & 86.50 & 4.69 & 77.59 & 67.30 & 80.75 & 5.92 \\ 
        & ST \cite{yang2022st} & CVPR (2022) & 85.46 & 77.70 & 91.25 & \textbf{3.75} & 82.23 & 73.53 & 89.25 & \textbf{4.16} & 72.54 & 62.69 & 74.0 & 4.92 \\
        & \textbf{CSDT} & \textbf{Ours} & \textbf{86.76} & \textbf{78.82} & \textbf{93.50} & 4.34 & \textbf{84.82} & \textbf{76.57} & 92.0 & 4.55 & \textbf{81.63} & \textbf{71.84} & \textbf{88.50} & 5.75 \\

        \midrule       
        \midrule
    \end{tabular}
    }
\end{table*}

Tables \ref{table:table2} and \ref{table:table3} present the SSTD performance of different SOTA SSL methods across three network configurations under noise affected by sunlight, earthlight, moonlight, and mixedlight noise. Performance metrics such as the Dice coefficient, mIoU, and \(P_d\) are expressed in percentage (\%) form, while the \(F_a\) is presented in an expanded \(10^{-4}\) format. The top performance metrics are highlighted in bold to signify the best results achieved. We conducted comparative experiments on the AstroStripeSet, utilizing only 1/4, 1/8, and 1/16 of the training set as labeled images, with the remainder being unlabeled images. The label `Sup.only` indicates training only with the labeled images. Compared to the `Sup.only` method, all evaluated SSL methods enhance SSTD performance by using new knowledge from the unlabeled images about stripe-like target distributions, particularly at lower labeling rates. These improvements not only affirm the effectiveness of the SSL approaches in SSTD but also underscore the method’s practical applicability. Our evaluation includes both quantitative comparisons and visual effect assessments.

\begin{table*}[t!]
    \centering
    \caption{The proposed CSDT architecture is compared with SOTA SSL methods across four types of stray light. Each type's performance is assessed using average Dice $\uparrow$(\%), mIoU $\uparrow$(\%), $P_d$ $\uparrow$(\%), and $F_a$ $\downarrow$($×10^{-4}$) metrics at labeling ratios of 1/4, 1/8, and 1/16.}
    \label{table:table3}
    \renewcommand{\arraystretch}{1.4}
    \resizebox{1.0\textwidth}{!}{
    \begin{tabular}{c| c| c|c c c c| c c c c |c c c c |c c c c} 
        \midrule
        \multirow{2}{*}{\textbf{Network}} & \multirow{2}{*}{\textbf{Method}} & \multirow{2}{*}{\textbf{Source}} & \multicolumn{4}{c|}{\textbf{Sun Light}} & \multicolumn{4}{c|}{\textbf{Earth Light}} & \multicolumn{4}{c|}{\textbf{Moon Light}} & \multicolumn{4}{c}{\textbf{Mixed Light}} \\  \cline{4-19}
        & & & \textbf{Dice} & \textbf{mIoU} & 
        \textbf{P}\textsubscript{\textbf{d}} & \textbf{F}\textsubscript{\textbf{a}} & 
        \textbf{Dice} & \textbf{mIoU} & 
        \textbf{P}\textsubscript{\textbf{d}} & \textbf{F}\textsubscript{\textbf{a}} & 
        \textbf{Dice} & \textbf{mIoU} & 
        \textbf{P}\textsubscript{\textbf{d}} & \textbf{F}\textsubscript{\textbf{a}} & 
        \textbf{Dice} & \textbf{mIoU} & 
        \textbf{P}\textsubscript{\textbf{d}} & \textbf{F}\textsubscript{\textbf{a}} \\
        \midrule
        \midrule
        \multirow{7}{*}{\makecell{\textbf{UNet} \\ \cite{ronneberger2015u}}} 
        & Sup.only \cite{ronneberger2015u} & MICCAI (2015) & 66.81 & 57.71 & 64.67 & 2.25 & 71.26 & 62.11 & 73.0 & 2.88 & 71.47 & 62.83 & 72.67 & 3.52 & 66.83 & 57.34 & 65.0 & 3.53\\
        & MT \cite{tarvainen2017mean} & Neurips (2017) & 75.88 & 66.53 & 76.67 & 2.77 & 79.51 & 70.31 & 82.67 & 3.25 & 78.43 & 69.76 & 81.0 & 3.74 & 74.62 & 65.43 & 73.67 & 2.68\\
        & UT \cite{liu2021unbiased} & ICLR (2021) & 78.53 & 69.16 & 82.67 & 5.79 & 80.63 & 71.39 & 86.67 & 6.25 & 80.33 & 71.23 & 87.0 & 6.54 & 74.18 & 64.74 & 75.33 & 15.35\\
        & ISMT \cite{yang2021interactive} & CVPR (2021) & 79.14 & 69.67 & 83.0 & 6.16 & 81.56 & 72.64 & 87.33 & 6.15 & 78.85 & 69.85 & 83.33 & 6.84 & 75.19 & 65.97 & 77.0 & 16.09 \\
        & PLMT \cite{mao2023semi} & ICASSP (2023) & 70.29 & 60.87 & 69.67 & 3.10 & 75.25 & 65.51 & 77.33 & 3.94 & 73.33 & 64.48 & 75.0 & 4.08 & 69.98 & 60.21 & 68.33 & 4.23\\
        & ST \cite{yang2022st} & CVPR (2022) & 70.80 & 61.51 & 69.67 & \textbf{1.93} & 76.50 & 67.10 & 78.0 & \textbf{2.64} & 75.13 & 66.12 & 75.67 & \textbf{2.99} & 72.27 & 62.53 & 71.33 & \textbf{2.41} \\
        & \textbf{CSDT} & \textbf{Ours} & \textbf{80.39} & \textbf{71.60} & \textbf{86.33} & 3.09 & \textbf{82.41} & \textbf{73.63} & \textbf{88.67} & 3.87 & \textbf{80.66} & \textbf{72.32} & \textbf{87.33} & 4.14 & \textbf{78.44} & \textbf{69.39} & \textbf{82.00} & 2.88 \\
        \midrule
        \midrule
        \multirow{7}{*}{\makecell{\textbf{UCTransNet} \\ \cite{wang2022uctransnet}}}
        & Sup.only \cite{wang2022uctransnet} & AAAI (2022) & 69.68 & 60.87 & 72.67 & 2.83 & 74.05 & 64.85 & 77.0 & 3.52 & 74.04 & 65.26 & 79.0 & 4.45 & 70.45 & 60.94 & 70.0 & 4.15\\
        & MT \cite{tarvainen2017mean} & Neurips (2017) & 75.92 & 66.16 & 78.67 & 3.75 & 80.11 & 70.59 & 83.33 & 4.39 & 78.38 & 69.01 & 82.67 & 5.09 & 76.60 & 66.97 & 78.33 & 5.99 \\
        & UT \cite{liu2021unbiased} & ICLR (2021) & 81.44 & 71.58 & 88.0 & 5.70 & 82.40 & 72.72 & 88.67 & 6.55 & 81.24 & 71.80 & 88.0 & 6.44 & 79.51 & 69.54 & 82.67 & 7.42\\
        & ISMT \cite{yang2021interactive} & CVPR (2021) & 80.43 & 70.52 & 86.67 & 4.05 & 81.83 & 72.42 & 89.0 & 4.74 & 81.19 & 71.40 & 86.33 & 5.38 & 78.50 & 68.70 & 83.0 & 4.01\\
        & PLMT \cite{mao2023semi} & ICASSP (2023) & 71.78 & 62.54 & 76.33 & 3.29 & 74.56 & 64.97 & 78.0 & 4.18 & 74.94 & 65.85 & 79.33 & 4.49 & 71.77 & 62.20 & 73.33 & 4.59\\
        & ST \cite{yang2022st} & CVPR (2022) & 69.92 & 61.08 & 72.33 & \textbf{2.03} & 75.15 & 65.70 & 78.67 & \textbf{2.95} & 73.25 & 64.55 & 76.67 & \textbf{3.35} & 69.94 & 60.58 & 68.33 & \textbf{2.30}\\
        & \textbf{CSDT} & \textbf{Ours} & \textbf{82.61} & \textbf{73.32} & \textbf{90.0} & 3.87 & \textbf{84.21} & \textbf{74.89} & \textbf{90.67} & 4.46 & \textbf{82.86} & \textbf{73.59} & \textbf{90.0} & 5.11 & \textbf{80.73} & \textbf{70.70} & \textbf{84.0} & 3.71 \\
        \midrule
        \midrule
        \multirow{7}{*}{\textbf{\makecell{MSSA-Net \\ (Ours)}}} 
        & Sup.only & \textbf{Ours} & 77.0 & 68.28 & 82.33 & 4.12 & 79.38 & 70.59 & 85.0 & 4.74 & 78.94 & 70.54 & 85.33 & 5.12 & 75.31 & 66.40 & 77.67 & 8.72 \\  
        & MT \cite{tarvainen2017mean} & Neurips (2017) & 78.94 & 70.01 & 82.67 & 3.99 & 82.14 & 73.58 & 87.33 & 4.84 & 81.63 & 73.14 & 86.67 & 4.86 & 76.34 & 67.65 & 78.67 & \textbf{3.79}\\ 
        & UT \cite{liu2021unbiased} & ICLR (2021) & 84.05 & 75.30 & 91.67 & 5.77 & 84.19 & 75.71 & 91.67 & 7.26 & 82.48 & 74.07 & 90.67 & 6.65 & 78.78 & 70.06 & 83.33 & 14.99\\ 
        & ISMT \cite{yang2021interactive}& CVPR (2021)  & 82.47 & 73.48 & 89.0 & 5.42 & 83.86 & 75.60 & 91.33 & 6.26 & 81.61 & 73.74 & 88.67 & 5.82 & 79.17 & 70.36 & 83.33 & 9.54\\ 
        & PLMT \cite{mao2023semi} & ICASSP (2023) & 80.11 & 70.65 & 84.67 & 4.69 & 83.28 & 74.28 & 89.33 & 5.88 & 80.0 & 71.29 & 86.67 & 5.40 &  79.16 & 70.30 & 84.67 & 4.75\\ 
        & ST \cite{yang2022st} & CVPR (2022) & 78.61 & 69.73 & 82.67 & \textbf{3.14} & 82.26 & 73.46 & 88.0 & \textbf{4.01} & 81.80 & 73.40 & 87.67 & \textbf{4.21} & 77.63 & 68.65 & 81.0 & 5.74\\
        & \textbf{CSDT} & \textbf{Ours} & \textbf{84.99} & \textbf{76.04} & \textbf{92.33} & 4.12 & \textbf{86.12} & \textbf{77.67} & \textbf{93.33} & 5.06 & \textbf{84.62} & \textbf{76.07} & \textbf{91.33} & 5.14 & \textbf{81.89} & \textbf{73.19} & \textbf{88.33} & 5.20 \\
        \midrule
        \midrule
    \end{tabular}
    }
\end{table*}

\begin{figure*}[h!]
  \centering
  \includegraphics[width=0.95\textwidth]{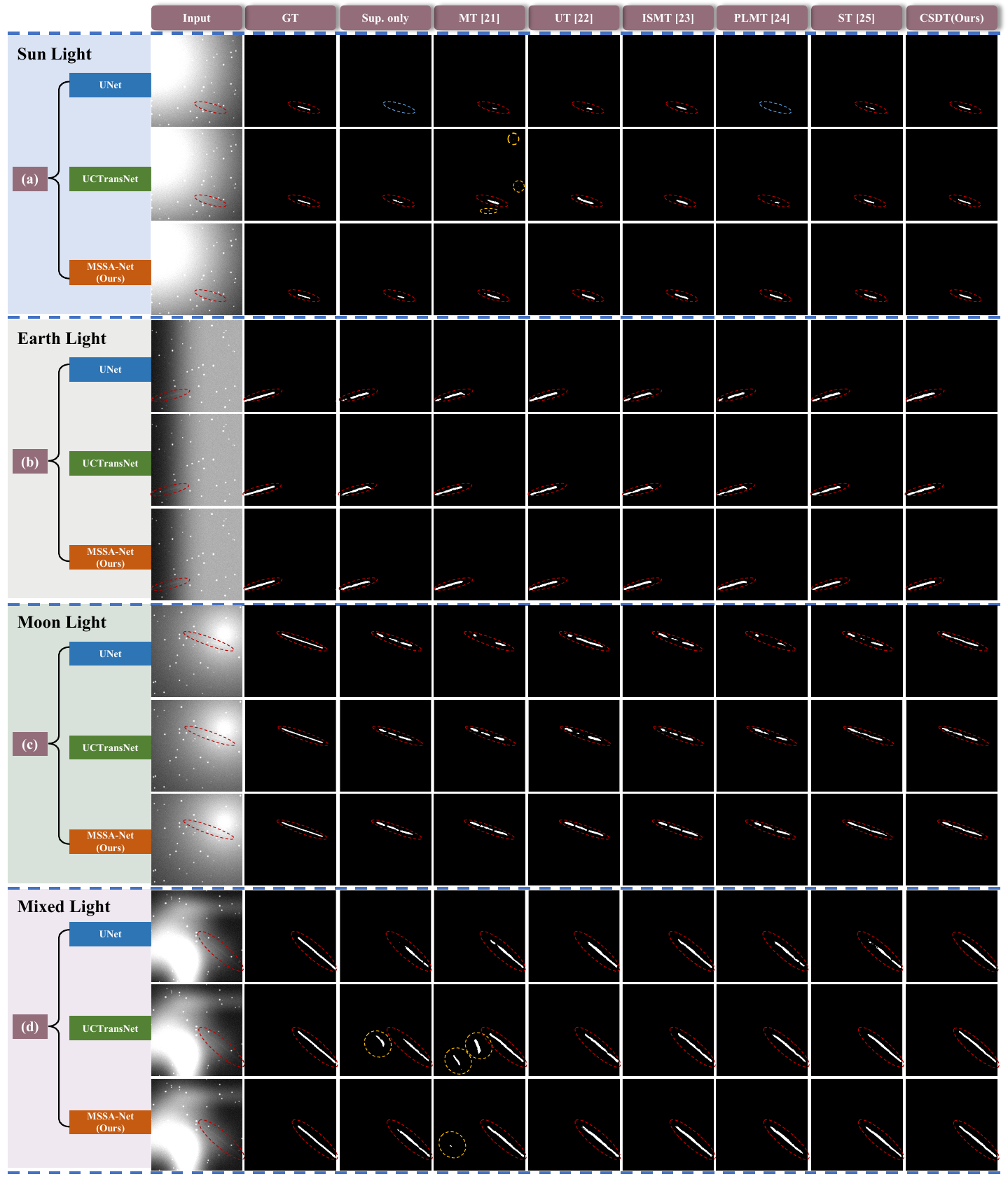}
  \caption{Visualization results of different SSL methods on AstroStripeSet using three network configurations: UNet, UCTransNet, and MSSA-Net as segmentation networks. Evaluations are conducted across these networks at a labeling rate of 1/4. (a) Visualization of detection results under sunlight interference. (b) Visualization of detection results under earth light interference. (c) Visualization of detection results under moonlight interference. (d) Visualization of detection results under mixed light interference. (Detected real stripe-like targets are highlighted in the dotted \textcolor{red}{red} circle, missed detections are marked in \textcolor{blue}{blue}, and false alarms are presented in \textcolor{yellow}{yellow}.)}
  \label{fig:fig5}
\end{figure*}

\subsubsection{Quantitative Comparison}
Tables \ref{table:table2} and \ref{table:table3} quantitatively present the average performance on the AstroStripeSet across various labeling rates (1/4, 1/8, 1/16). Our comparative experiments employed UNet, UCTransNet, and our proposed MSSA-Net as segmentation networks. Under these network configurations, our CSDT architecture all achieved notable success. While our CSDT did not achieve the best \(F_a\), it excelled in other key performance metrics, surpassing existing SOTA SSL methods in the Dice coefficient, mIoU, and \(P_d\). 

In Tables \ref{table:table2}, under the UNet configuration with a labeling ratio of 1/4, our CSDT method achieved a Dice coefficient of 83.35\%, a mIoU of 75.72\%, and a \(P_d\) of 89.75\%. These results represent improvements of 6.28\%, 6.90\%, and 10.0\% over the 'Sup.only' method, and also surpass the next best performing ISMT method by 0.08\%, 0.86\%, and 1.50\%, respectively.

When our MSSA-Net operated as the teacher-student network, all SSL methods showed more significant detection performance improvements compared to both the UNet and UCTransNet configurations. Specifically, with a labeling ratio of 1/8, SSL methods, including MT, UT, ISMT, PLMT, ST, and CSDT, showed the following increases in \(P_d\): under MSSA-Net, improvements over UNet were 6.50\%, 7.50\%, 2.25\%, 12.75\%, 14.50\%, and 5.50\%, respectively; while improvements over UCTransNet were 5.75\%, 1.75\%, 1.25\%, 5.50\%, 10.25\%, and 1.25\%, respectively. Additionally, at a labeling rate of 1/8, the CSDT method with MSSA-Net even outperformed the CSDT method with UNet at a 1/4 labeling rate. It also matched the performance of the CSDT method with UCTransNet at a 1/4 labeling rate. Similar trends were also observed in the UT and PLMT methods using different networks. This phenomenon highlights the significant advantage of MSSA-Net, especially in scenarios with limited labeled images.

Table \ref{table:table3} presents experiments on AstroStripeSet with varying labeling ratios (1/4, 1/8, 1/16) for each type of stray light noise, showing the average detection performance for each category of stray light at these ratios. Our proposed CSDT architecture consistently outperforms other SSL methods across three network configurations. Moreover, compared to the Sup.only method, all SSL methods significant improve in Dice, mIoU, and \(P_d\). However, their ability to suppress false alarms has decreased, possibly due to pseudo-label filtering during training of unlabeled images inevitably includes non-target false alarm sources, leading to a higher rate of \(F_a\). This limitation highlights a persistent challenge for SSL methods. Consequently, based on our APL mechanism, further suppressing false alarm sources in pseudo-labels and adding a rejection strategy to minimize learning of non-target features constitutes the focus of our future research. 

To further highlight the advantages of our CSDT framework, we refer to performance indices from the subset tests involving earth light noise. Under MSSA-Net configuration, the CSDT shows improvements of 6.74\%, 7.08\%, and 8.33\% in Dice, mIoU, and \(P_d\) respectively over the Sup.only method; and 1.93\%, 1.96\%, and 1.66\% improvements over the second-ranked UT method. These improvements are largely attributed to the collaborative efforts of the static and dynamic teachers, along  with the APL and EMA mechanisms. These settings enable the model to continuously refine the quality of pseudo-labels throughout the training process, fostering a mutually beneficial and progressive training environment for both the DT and S models. 

In contrast, the ST method, which utilized a fixed threshold for pseudo-label selection based on model prediction confidence, frequently included poor-quality pseudo-labels in the training process. This inclusion of less accurate labels could limit the model's overall accuracy. Although the MT method dynamically updated the teacher's weights via EMA, its feedback strategy was fixed, just focusing on maintaining consistency between the teacher and student predictions. This could lead to the model overfit noisy labels, reducing its ability to deeply understand real stripe-like patterns. Similarly, SSL methods like UT, ISMT, and PLMT relied heavily on the pre-trained single-teacher models, leading to rigid teaching approaches that limit the model's ability to explore new stripe-like pattern distributions. Furthermore, these pre-trained teachers may prematurely adapt to the stripe-like pattern distribution during EMA updates, leading to overfitting and limiting their adaptability in the noisy and variable space scenarios.

\begin{figure*}[h!]
  \centering
  \includegraphics[width=0.98\textwidth]{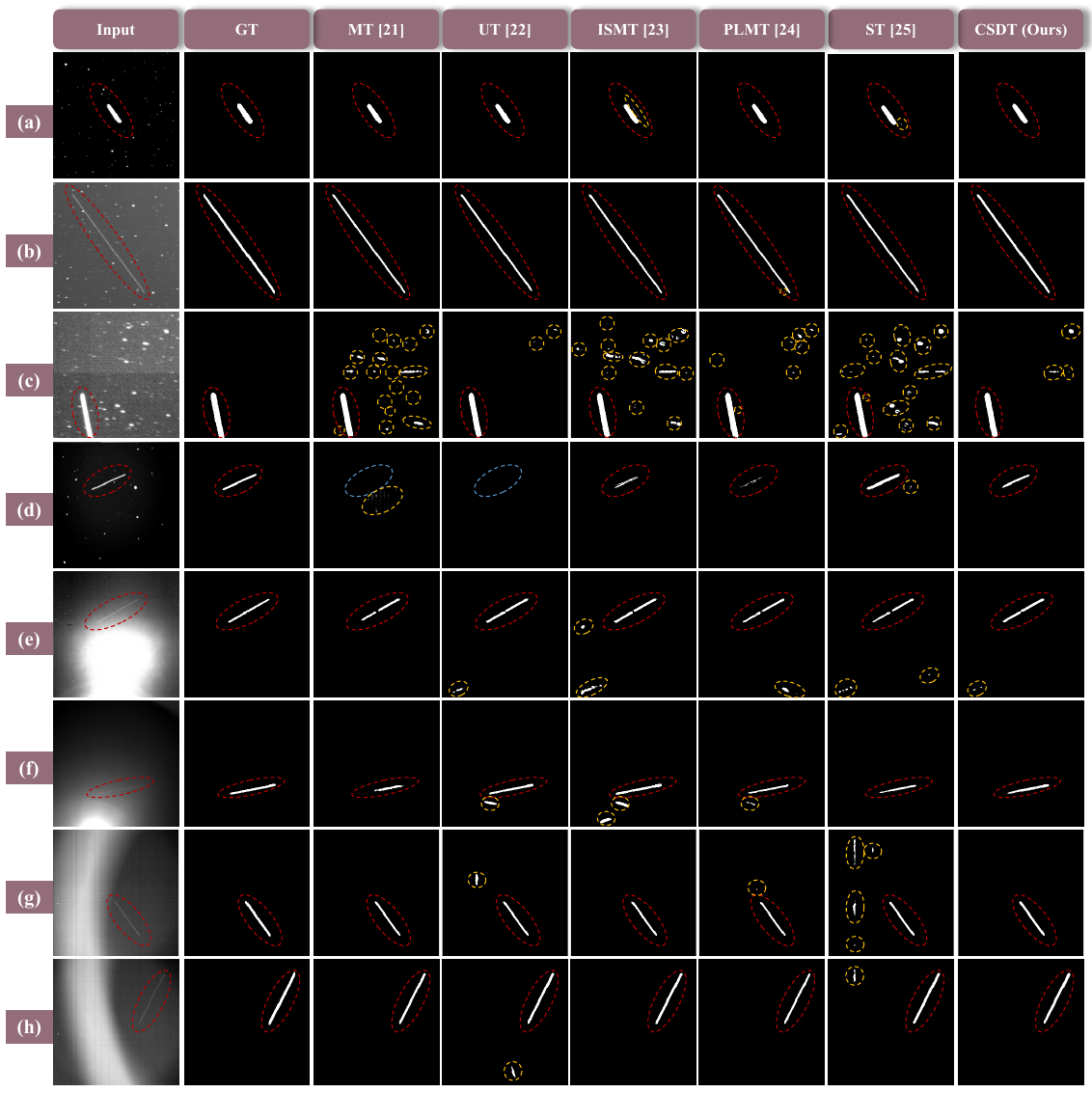}
  \caption{Visualization of the SSTD results using various SSL methods on real-world datasets with MSSA-Net as the teacher-student network at a 1/16 labeling rate. (a) shows data from \cite{lin2021new}. (b) presents data from the \cite{li2022bsc}. (c) displays data from \cite{liu2023multi}. (d) features data collected on the Internet. (e) and (f) are real background images collected by our on-orbit space-based cameras. (g) and (h) are real background images collected by our ground-based telescopes. (Detected real stripe-like targets are highlighted in the dotted \textcolor{red}{red} circle, missed detections are marked in \textcolor{blue}{blue}, and false alarms are presented in \textcolor{yellow}{yellow}.)}
  \label{fig:fig6}
\end{figure*}

\subsubsection{Visual Effect Assessments}

Alongside quantitative assessments, we provided visual comparisons with other SSL methods at a labeling rate of 1/4. Fig. \ref{fig:fig5} showcases the visual outcomes of different SSL methods under three segmentation networks. Figs.~\hyperref[fig:fig5]{\ref*{fig:fig5}(a)--(d)} show the detection performance of faint stripe-like space targets under sunlight, earthlight, moonlight, and mixedlight, respectively. Areas of detected real targets are highlighted in the dotted red circle, missed detections are marked in blue, and false alarms are shown in yellow. It is evident that the proposed CSDT framework significantly enhances SSTD capabilities across all three segmentation networks. It preserves target integrity, minimizes the impact of background noise, and improves overall segmentation accuracy and robustness. 

Specifically, Fig.~\hyperref[fig:fig5]{\ref*{fig:fig5}(a)} illustrates that using UNet as the teacher-student network, both the Sup.only and PLMT approachs fail to detect the faint stripe-like target. This detection capability is significantly enhanced by SSL methods such as ST, UT, MT, ISMT, and particularly by the proposed CSDT architecture, showcasing the effectiveness of SSL methods in the SSTD task. In Figs.~\hyperref[fig:fig5]{\ref*{fig:fig5}(b)-(c)}, when using UNet and UCTransNet as the teacher-student networks, almost all compared SSL methods show target breakages in their visual results, except the proposed CSDT architecture. This clearly highlights the significant advantage of the dual-teacher adaptive collaborative teaching strategy in our CSDT framework. When MSSA-Net serves as the teacher-student network, it significantly reduces the issue of stripe-like target area breakage in all SSL methods, especially within the UT, ISMT, and PLMT methods, leading to notable improvements. These improvements underscore the
superiority of MSSA-Net in preserving the integrity of stripe-like targets, even in the scenarios of low SNR. Moreover, as shown in Fig.~\hyperref[fig:fig5]{\ref*{fig:fig5}(d)}, when using MSSA-Net as the teacher-student network, each SSL method more effectively maintains target integrity compared to UNet. MSSA-Net also more effectively suppresses false alarms than UCTransNet in all SSL settings, highlighting its strengths in enhancing stripe-like target continuity and reducing false detections in challenging space scenarios. The series of visual comparisons in the Fig. \ref{fig:fig5} highlight the exceptional performance of MSSA-Net in handling low SNR targets across various scenarios. These comparisons also demonstrate that the CSDT framework effectively extracts new knowledge from unlabeled space images, thereby improving the model's generalization performance.

\begin{table*}[ht]
\centering
\caption{Ablation study of assessing zero-shot generalization capabilities on real-world datasets using MSSA-Net, trained under various SSL methods on the AstroStripeSet at a 1/16 labeling rate, and evaluated by mIoU $\uparrow$(\%).}
\label{table:table9}
\renewcommand{\arraystretch}{1.4}
\resizebox{\textwidth}{!}{  
\begin{tabular}{c|c|c|c|c|c|c|c|c|c|c}
    \toprule
    \multirow{2}{*}{\textbf{Network}} & \multirow{2}{*}{\textbf{Method}} & \multirow{2}{*}{\textbf{Source}} & \multicolumn{4}{c|}{Others} & \multicolumn{4}{c}{Ours}   \\
    \cline{4-11}
    & & & Img (a) & Img (b) & Img (c) & Img (d) & Img (e) & Img (f) & Img (g) & Img (h)   \\ 
    \midrule
    \midrule
    \multirow{6}{*}{\textbf{\makecell{MSSA-Net \\ (Ours)}}} 
    & MT \cite{tarvainen2017mean} & NeurIPS (2017) & 84.23 & 76.36 & 61.85 & 0.0 & 71.93 & 51.83 & 78.26 & 82.48  \\
    & UT \cite{liu2021unbiased} & ICLR (2021) & 89.0 & 78.87 & \textbf{91.93} & 0.0 & 74.41 & 69.68 & 64.31 & 77.02  \\
    & ISMT \cite{yang2021interactive} & CVPR (2021) & 85.16 & 78.25 & 54.19 & 42.22 & 54.31 & 54.46 & 79.86 & 87.16 \\
    & PLMT \cite{mao2023semi} & ICASSP (2023) & 87.14 & 75.46 & 85.17 & 10.09  & 71.60 & 63.85 & 73.02 & 74.77 \\
    & ST \cite{yang2022st} & CVPR (2022) & 87.65 & 78.95 & 53.63 & 68.40 & 70.29 & 60.37 & 52.25 & 79.62 \\
    & \textbf{CSDT} & \textbf{Ours} & \textbf{90.25} & \textbf{91.77} & 84.01 & \textbf{73.09} & \textbf{75.05} & \textbf{77.97} & \textbf{90.02} & \textbf{87.57} \\
    \bottomrule
    \bottomrule
\end{tabular}
}
\end{table*}

\subsection{Ablation Study}
In this section, we explored the effectiveness of our proposed framework through comprehensive ablation studies. We assessed the CSDT framework's generalization capabilities with real-world datasets in zero-shot settings; the contributions of each MSSA-Net component; the role of dual teachers; the impact of the APL strategy; and the role of each loss function.

\subsubsection{Zero-Shot Generalization Capabilities}
In this section, we specifically conducted zero-shot generalization verification experiments to assess the model's performance on unseen real-world scenarios from various sources. To the best of our knowledge, this aspect is often overlooked in most papers within the SSTD field. Fig. \ref{fig:fig6} displays the zero-shot generalization results of different SSL methods on real-world image datasets using MSSA-Net as the teacher-student network at a labeling rate of 1/16. Fig.~\hyperref[fig:fig6]{\ref*{fig:fig6}(a)} features data from \cite{lin2021new}; Fig.~\hyperref[fig:fig6]{\ref*{fig:fig6}(b)} uses data from \cite{li2022bsc}; Fig.~\hyperref[fig:fig6]{\ref*{fig:fig6}(c)} incorporates data from \cite{liu2023multi}; and Fig.~\hyperref[fig:fig6]{\ref*{fig:fig6}(d)} includes data collected from the Internet; Figs.~\hyperref[fig:fig6]{\ref*{fig:fig6}(e)--(f)} and Figs.~\hyperref[fig:fig6]{\ref*{fig:fig6}(g)--(h)} show real-space background images taken by our on-orbit space-based camera and ground-based telescope, respectively. These figures reveal significant differences in zero-shot generalization performance among the SSL methods.

As demonstrated in Table \ref{table:table9}, the CSDT framework delivers the best detection results overall, except for img(c), where UT and PLMT methods perform slightly better. This discrepancy is due to the APL strategy incorporating many high-quality stripe-like pseudo-labels, which enhances stripe-like pattern segmentation but might also lead to some false alarms in the presence of stripe-like noise. For other SSL methods, stripe-like targets are detected in simpler background scenarios (Figs.~\hyperref[fig:fig6]{\ref*{fig:fig6}(a)--(b)}). However, false detections (yellow marks, Fig.~\hyperref[fig:fig6]{\ref*{fig:fig6}(c)}, Fig.~\hyperref[fig:fig6]{\ref*{fig:fig6}(f)}, Fig.~\hyperref[fig:fig6]{\ref*{fig:fig6}(g)}) and missed detections (blue marks, Fig.~\hyperref[fig:fig6]{\ref*{fig:fig6}(d)}) occurred in more challenging cases. Additionally, in Fig.~\hyperref[fig:fig6]{\ref*{fig:fig6}(e)}, SSL methods like ST, MT, and PLMT exhibit significant breakage in the stripe-like target areas, and UT and ISMT also result in more false alarms than the proposed CSDT method. These outcomes highlight the limitations of these rigid teaching models, which cannot dynamically adapt their strategies to improve pseudo-label quality, thus restricting improvements in generalization capabilities to unknown data. Thus, we not only validate the effectiveness of our proposed method in a standard test environment but also demonstrate its reliability in real-world applications.

\subsubsection{MSSA-Net Components}

In this section, we evaluated the effectiveness of the MDPC and FMWA blocks, two critical components proposed in MSSA-Net. We utilized 1/4 of the labeled space images from AstroStripeSet for Sup.only training, and evaluate the contribution of each block.  As detailed in Table \ref{table:table4}, the Baseline achieved the lowest $F_a$ compared to MSSA-Net. This is because the MDPC and FMWA blocks, while enhancing weakly textured stripe-like target feature extraction, also capture some stray light regions similar to stripe-like targets, leading to a slight increase in $F_a$. However, MSSA-Net achieved significant improvements in other metrics, with increases of 4.67\% in mIoU and 8.0\% in $P_d$ compared to network with only FMWA blocks; 1.17\% and 1.75\% over network with only MDPC blocks; and 7.94\% and 12.50\% over the baseline configuration. These results demonstrate that the multi-receptive field stripe-like target feature extraction mechanism of the MDPC blocks, combined with the enhancement provided by the FMWA blocks, leads to significant improvements in the SSTD performance.

\begin{table}[h!]
    \centering
    \caption{Ablation study of the MDPC and FMWA blocks in the MSSA-Net on the AstroStripeSet at a 1/4 labeling rate, evaluated by Dice $\uparrow$(\%), mIoU  $\uparrow$(\%), $P_d$ $\uparrow$(\%), and $F_a$ $\downarrow$($×10^{-4}$).}
    \label{table:table4}
    \small
    \renewcommand{\arraystretch}{1.4}
    \begin{tabularx}{\columnwidth}{c |c c| c c c c} 
        \midrule
        \multirow{2}{*}{\textbf{Network}} & \multicolumn{2}{c|}{\textbf{Module}} & \multicolumn{4}{c}{\textbf{Average Metrics}}  \\  
        \cline{2-7} 
        & \textbf{MDPC} & \textbf{FMWA} & \textbf{Dice}  &
        \textbf{mIoU}  & 
        \textbf{P}\textsubscript{\textbf{d}}  & \textbf{F}\textsubscript{\textbf{a}}  \\
        \midrule
        \midrule
        \multirow{4}{*}{\textbf{Baseline}} 
        & \checkmark & \checkmark & \textbf{84.98} & \textbf{76.76} & \textbf{92.25} & 4.61\\
        & $\times$ & \checkmark & 80.17 & 72.09 & 84.25 & 5.34  \\
        & \checkmark & $\times$ & 83.50 & 75.59 & 90.50 & 5.13 \\
        & $\times$ & $\times$ &  77.07 & 68.82 & 79.75 & \textbf{3.44} \\
        \midrule
        \midrule
    \end{tabularx}
\end{table}

Furthermore, we conducted an ablation study to determine the optimal number of MDPC blocks. Table \ref{table:table5} reveal that the best detection performance is achieved with three MDPC blocks. 
This finding indicates that three MDPC blocks strike an optimal balance between coverage and specificity, effectively preserving pixel-level details of stripe-like targets while enhancing detection accuracy at the target-level. This balance is crucial in complex space scenarios, where too few blocks might miss finer stripe-like target details and too many could lead to overfitting on noise. 

\begin{table}[h!]
    \centering
    \caption{Ablation study of the numbers of MDPC blocks in MSSA-Net on the AstroStripeSet at a 1/4 labeling rate, evaluated by Dice $\uparrow$(\%), mIoU  $\uparrow$(\%), $P_d$ $\uparrow$(\%), and $F_a$ $\downarrow$($×10^{-4}$).}
    \label{table:table5}
    \small
    \renewcommand{\arraystretch}{1.4}
    \begin{tabularx}{\columnwidth}{c |c | Y Y Y Y} 
        \midrule
        \multirow{2}{*}{\textbf{Module}} & \multirow{2}{*}{\textbf{Numbers}} & \multicolumn{4}{c}{\textbf{Average Metrics}}  \\  
        \cline{3-6} 
        &  &  \textbf{Dice}  & \textbf{mIoU}  & 
        \textbf{P}\textsubscript{\textbf{d}}  & \textbf{F}\textsubscript{\textbf{a}}  \\
        \midrule
        \midrule
        \multirow{4}{*}{\textbf{MDPC}} 
        & 1 & 81.96 & 74.30 & 88.25 & 4.85\\
        & 2  & 83.82 & 76.10 & 91.50 & 4.83 \\
        & \textbf{3}  &\textbf{84.98}  & \textbf{76.76} & \textbf{92.25} & \textbf{4.61} \\
        & 4  &  82.50 & 74.65 & 89.50 & 4.77 \\
        \midrule
        \midrule
    \end{tabularx}
\end{table}

\subsubsection{Single-Teacher vs. Dual-Teacher Supervision}
Table \ref{table:table6} displays the SSTD performance at a labeling rate of 1/16 across various configurations: using both teachers (ST and DT together), one teacher (ST or DT alone), and no teacher (Sup.only). With only the ST model, the SSL framework utilizes pseudo-label segmentation loss $\mathcal{L}_u$ for unlabeled images, without involving EMA. With only the DT model, it depends on consistency loss $\mathcal{L}_c$ to use unlabeled images, aiming to maintain output consistency between DT and the S model. Table \ref{table:table6} shows that our CSDT framework, which integrates both the ST and DT models, significantly outperforms other setups. This enhanced performance is attributed to the fusion of the strengths of the ST and DT models, as well as the application of the APL strategy. It adaptively employs high-quality pseudo-labels, which not only enhances the S model’s performance but also continuously improves DT’s performance through the EMA mechanism, thereby further refining the pseudo-label quality and creating a positive feedback loop. This strategy efficiently utilizes unlabeled images and minimizes the impact of incorrect labels, thus boosting overall accuracy and model generalization across diverse space scenarios. Ablation study results further confirm that CSDT architecture is highly effective for the SSTD task.

\begin{table}[h!]
    \centering
    \caption{Ablation study of comparing single-teacher and dual-teacher setups on the AstroStripeSet at a 1/16 labeling rate, assessed by Dice $\uparrow$(\%), mIoU  $\uparrow$(\%), $P_d$ $\uparrow$(\%), and $F_a$ $\downarrow$($×10^{-4}$).}
    \label{table:table6}
    \small
    \renewcommand{\arraystretch}{1.4}
    \begin{tabularx}{\columnwidth}{c |c c| Y Y Y Y} 
        \midrule
        \multirow{2}{*}{\textbf{Network}} & \multicolumn{2}{c|}{\textbf{Teacher Type}} & \multicolumn{4}{c}{\textbf{Average Metrics}}  \\  
        \cline{2-7} 
        & \textbf{DT} & \textbf{ST} & \textbf{Dice}  & \textbf{mIoU}  & 
        \textbf{P}\textsubscript{\textbf{d}} & \textbf{F}\textsubscript{\textbf{a}} \\
        \midrule
        \midrule
        \multirow{4}{*}{\textbf{\makecell{MSSA-Net \\ (Ours)}}} 
        & \checkmark & \checkmark & \textbf{81.63}& \textbf{71.84} & \textbf{88.50} & \textbf{5.75} \\
        & $\times$ & \checkmark & 75.49 & 65.55 & 75.50 & 6.27 \\
        & \checkmark & $\times$ & 76.59 & 67.16 & 81.25 & 6.07 \\
        & $\times$ & $\times$ & 69.69 & 60.09 & 71.25 & 7.93 \\
        \midrule
        \midrule
    \end{tabularx}
\end{table}

\subsubsection{Evaluation of APL Strategy}

Within the CSDT framework, we assessed the average SSTD performance using different pseudo-labeling (PL) strategies at a labeling rate of 1/16. Since the ST model is pre-trained, it initially outperforms the DT model. However, DT's performance enhances as it progressively learns from the S model. Therefore, it is obvious that we can directly use the prediction of the ST model for the initial $N$ epochs and switch to DT’s predictions thereafter. We denote this method as the ST $\to$ DT PL strategy. Additionally, using either the intersection (ST $\cap$ DT) or the union (ST $\cup$ DT) of the predictions from ST and DT as pseudo-labels represent two other intuitive PL strategies. To demonstrate the superiority of our APL strategy, we compared it against these more straightforward PL approaches that lack adaptive selection mechanisms.
As illustrated in Table \ref{table:table7}, the APL strategy significantly outperforms the fixed PL strategies across all metrics. This is because the APL strategy cleverly utilizes the prior knowledge of directional consistency inherent in the stripe-like space targets during the pseudo-labeling process. Based on this, we introduce a novel metric of 'linearity' enabling the adaptive selection of optimal pseudo-labels from the ST and DT models, thereby better teaching the S model during training.

\begin{table}[h!]
    \centering
    \caption{Ablation study of various PL strategies on the AstroStripeSet at a 1/16 labeling rate, measured by Dice $\uparrow$(\%), mIoU  $\uparrow$(\%), \\ $P_d$ $\uparrow$(\%), and $F_a$ $\downarrow$($×10^{-4}$).}
    \label{table:table7}
    \small
    \resizebox{1.0\columnwidth}{!}{
    \renewcommand{\arraystretch}{1.4}
    \begin{tabular}{c |c|c| c c c c } 
        \midrule
        \multirow{2}{*}{\textbf{Network}} & \multirow{2}{*}{\textbf{PL strategy}} & \multirow{2}{*}{\textbf{Epochs}} &  \multicolumn{4}{c}{\textbf{Average Metrics}}  \\  
        \cline{4-7} 
        & & &   \textbf{Dice}  & \textbf{mIoU}  & 
        \textbf{P}\textsubscript{\textbf{d}}  & \textbf{F}\textsubscript{\textbf{a}}  \\
        \midrule
        \midrule
        \multirow{6}{*}{\textbf{\makecell{MSSA-Net \\ (Ours)}}} 
        & APL (Ours) & Overall & \textbf{81.63} & \textbf{71.84} & \textbf{88.50} & \textbf{5.75} \\
        & ST $\cap$ DT & Overall & 75.74 & 66.24 & 77.50 & 6.42 \\
        & ST $\cup$ DT & Overall & 76.86 & 67.29 & 80.0 & 6.95 \\
        & ST $\to$ DT & 30 & 80.67 & 71.02 & 86.5 & 7.94\\
        & ST $\to$ DT & 40 & 80.74 & 71.16 & 86.0 & 5.99 \\
        & ST $\to$ DT & 50 & 80.32 & 70.61 & 85.50 & 7.32 \\
        \midrule
        \midrule
    \end{tabular}
    }
\end{table}

\subsubsection{ Impact of Loss Functions}

Within the CSDT framework, we conducted a loss function ablation study using MSSA-Net as the teacher-student network at a 1/16 labeling rate. We explored three different loss functions for the S model: supervised loss \(L_s\) for labeled images, pseudo-label supervised loss \(L_u\) for unlabeled images, and consistency loss \(L_c\) for maintaining output consistency between the S model and DT model across different augmentations of unlabeled images. Table \ref{table:table8} shows the impact of these losses on the average performance under four types of stray light noise on the AstroStripeSet. 

\begin{table}[h!]
    \centering
    \caption{Ablation study of various loss function combinations on the AstroStripeSet at a 1/16 labeling rate, evaluated by \\ Dice $\uparrow$(\%), mIoU  $\uparrow$(\%), $P_d$ $\uparrow$(\%), and $F_a$ $\downarrow$($×10^{-4}$).}
    \label{table:table8}
    \small
    \renewcommand{\arraystretch}{1.4}
    \begin{tabularx}{\columnwidth}{c|c c c|Y Y Y Y}  
        \midrule
        \multirow{2}{*}{\textbf{Network}} & \multicolumn{3}{c|}{\textbf{Loss Function}} & \multicolumn{4}{c}{\textbf{Average Metrics}}  \\
        \cline{2-8}
        & \textbf{$L_s$} & \textbf{$L_u$} & \textbf{$L_c$} & \textbf{Dice}  & \textbf{mIoU}  & \textbf{P}\textsubscript{\textbf{d}}  & \textbf{F}\textsubscript{\textbf{a}}  \\  
        \midrule
        \midrule
        \multirow{4}{*}{\textbf{\makecell{MSSA-Net \\ (Ours)}}} &
        \checkmark & \checkmark & \checkmark & \textbf{81.63} &\textbf{71.84}  & \textbf{88.50} & \textbf{5.75} \\
        & \checkmark & \checkmark & $\times$ & 80.55 & 70.84 & 85.75 & 5.80 \\ 
        & \checkmark & $\times$ & \checkmark & 76.59 &67.16 & 81.25 & 6.07 \\
        & \checkmark & $\times$ & $\times$ & 69.69 & 60.09 & 71.25 & 7.93 \\ 
        \midrule
        \midrule
    \end{tabularx}
\end{table}

Table \ref{table:table8} shows that using \(L_s\), \(L_u\), and \(L_c\) together yields the best performance in Dice, mIoU, $P_d$, and $F_a$ metrics, with specific values of 81.63\%, 71.84\%, 88.50\%, and 5.75\% respectively. This confirms that integrating these three loss functions significantly enhances the accuracy of SSTD. Upon removing \(L_c\), a noticeable decline in performance occurs, with Dice and mIoU values decreasing to 80.55\% and 70.84\%, respectively. This underscores the vital role of consistency loss \(L_c\) in utilizing unlabeled images for training. The subsequent removal of \(L_u\) leads to a more pronounced performance drop, with Dice and mIoU plummeting to 76.59\% and 67.16\%. This illustrates the critical importance of pseudo-label loss \(L_u\) in enhancing the model's generalization capabilities and minimizing overfitting. Relying solely on \(L_s\) results in the lowest performance, with Dice and mIoU at 69.69\% and 60.09\%, respectively. Therefore, the pseudo-label supervised loss \(L_u\) is crucial for harnessing the latent information in unlabeled images, enabling the model to adapt to the intrinsic data variability and improve robustness. Consistency loss \(L_c\) ensures that the model's predictions are stable across different augmentations of the same input, which is vital for maintaining reliable predictions in diverse and complex space imaging scenarios.

\section{Conclusion}
\label{sec:conc}

This paper presents the innovative CSDT SSL framework for the SSTD task, addressing challenges like low signal-to-noise ratios and variability of stripe-like targets in space scenarios. The CSDT framework employs the proposed MSSA-Net as the network architecture for dual-teacher and student models. It integrates static and dynamic teacher models with a customized adaptive pseudo-labeling (APL) strategy and exponential moving average (EMA) mechanism, enhancing pseudo-label quality and boosting model generalization for the SSTD task. Our MSSA-Net, incorporating the designed MDPC and FMWA blocks, significantly improves the detection of stripe-like targets in various complex space scenarios compared to other segmentation networks. It excels across all SSL and sup.only training methods, performing particularly well within the proposed CSDT framework. The MSSA-Net within CSDT framework achieves state-of-the-art performance on the challenging AstroStripeSet and various ground-based and space-based real-world datasets. We believe that the CSDT framework can be effectively adapted to various remote sensing tasks by customizing different pseudo-labeling strategies. Looking ahead, we also plan to further optimize the proposed CSDT framework to explore its potential for more extensive applications.


%

\ifCLASSOPTIONcaptionsoff
  \newpage
\fi


\begin{thebibliography}{10}
\providecommand{\url}[1]{#1}
\csname url@samestyle\endcsname
\providecommand{\newblock}{\relax}
\providecommand{\bibinfo}[2]{#2}
\providecommand{\BIBentrySTDinterwordspacing}{\spaceskip=0pt\relax}
\providecommand{\BIBentryALTinterwordstretchfactor}{4}
\providecommand{\BIBentryALTinterwordspacing}{\spaceskip=\fontdimen2\font plus
\BIBentryALTinterwordstretchfactor\fontdimen3\font minus \fontdimen4\font\relax}
\providecommand{\BIBforeignlanguage}[2]{{%
\expandafter\ifx\csname l@#1\endcsname\relax
\typeout{** WARNING: IEEEtran.bst: No hyphenation pattern has been}%
\typeout{** loaded for the language `#1'. Using the pattern for}%
\typeout{** the default language instead.}%
\else
\language=\csname l@#1\endcsname
\fi
#2}}
\providecommand{\BIBdecl}{\relax}
\BIBdecl

\bibitem{li2020space}
Z.~Li, Y.~Wang, and W.~Zheng, ``Space-based optical observations on space debris via multipoint of view,'' \emph{International Journal of Aerospace Engineering}, vol. 2020, pp. 1--12, 2020.

\bibitem{wirnsberger2015space}
H.~Wirnsberger, O.~Baur, and G.~Kirchner, ``Space debris orbit prediction errors using bi-static laser observations. case study: Envisat,'' \emph{Advances in Space Research}, vol.~55, no.~11, pp. 2607--2615, 2015.

\bibitem{zhang2022dynamics}
J.~Zhang, A.~Shi, and K.~Yang, ``Dynamics of tethered-coulomb formation for debris deorbiting in geosynchronous orbit,'' \emph{Journal of Aerospace Engineering}, vol.~35, no.~3, p. 04022015, 2022.

\bibitem{liu2020topological}
D.~Liu, B.~Chen, T.-J. Chin, and M.~G. Rutten, ``Topological sweep for multi-target detection of geostationary space objects,'' \emph{IEEE Transactions on Signal Processing}, vol.~68, pp. 5166--5177, 2020.

\bibitem{diprima2018efficient}
F.~Diprima, F.~Santoni, F.~Piergentili, V.~Fortunato, C.~Abbattista, and L.~Amoruso, ``Efficient and automatic image reduction framework for space debris detection based on gpu technology,'' \emph{Acta Astronautica}, vol. 145, pp. 332--341, 2018.

\bibitem{liu2020space1}
D.~Liu, X.~Wang, Z.~Xu, Y.~Li, and W.~Liu, ``Space target extraction and detection for wide-field surveillance,'' \emph{Astronomy and Computing}, vol.~32, p. 100408, 2020.

\bibitem{yao2022adaptive}
Y.~Yao, J.~Zhu, Q.~Liu, Y.~Lu, and X.~Xu, ``An adaptive space target detection algorithm,'' \emph{IEEE Geoscience and Remote Sensing Letters}, vol.~19, pp. 1--5, 2022.

\bibitem{lin2021new}
B.~Lin, L.~Zhong, S.~Zhuge, X.~Yang, Y.~Yang, K.~Wang, and X.~Zhang, ``A new pattern for detection of streak-like space target from single optical images,'' \emph{IEEE Transactions on Geoscience and Remote Sensing}, vol.~60, pp. 1--13, 2021.

\bibitem{jiang2022automatic}
P.~Jiang, C.~Liu, W.~Yang, Z.~Kang, and Z.~Li, ``Automatic space debris extraction channel based on large field of view photoelectric detection system,'' \emph{Publications of the Astronomical Society of the Pacific}, vol. 134, no. 1032, p. 024503, 2022.

\bibitem{felt2024seeing}
V.~Felt and J.~Fletcher, ``Seeing stars: Learned star localization for narrow-field astrometry,'' in \emph{Proceedings of the IEEE/CVF Winter Conference on Applications of Computer Vision}, 2024, pp. 8297--8305.

\bibitem{lu2023fast}
K.~Lu, H.~Li, L.~Lin, R.~Zhao, E.~Liu, and R.~Zhao, ``A fast star-detection algorithm under stray-light interference,'' in \emph{Photonics}, vol.~10, no.~8.\hskip 1em plus 0.5em minus 0.4em\relax MDPI, 2023, p. 889.

\bibitem{xu2020stray}
Z.~Xu, D.~Liu, C.~Yan, and C.~Hu, ``Stray light nonuniform background correction for a wide-field surveillance system,'' \emph{Applied Optics}, vol.~59, no.~34, pp. 10\,719--10\,728, 2020.

\bibitem{li2022bsc}
Y.~Li, Z.~Niu, Q.~Sun, H.~Xiao, and H.~Li, ``Bsc-net: Background suppression algorithm for stray lights in star images,'' \emph{Remote Sensing}, vol.~14, no.~19, p. 4852, 2022.

\bibitem{liu2023multi}
L.~Liu, Z.~Niu, Y.~Li, and Q.~Sun, ``Multi-level convolutional network for ground-based star image enhancement,'' \emph{Remote Sensing}, vol.~15, no.~13, p. 3292, 2023.

\bibitem{hickson2018fast}
P.~Hickson, ``A fast algorithm for the detection of faint orbital debris tracks in optical images,'' \emph{Advances in Space Research}, vol.~62, no.~11, pp. 3078--3085, 2018.

\bibitem{levesque2007image}
M.~P. Levesque and S.~Buteau, ``Image processing technique for automatic detection of satellite streaks,'' \emph{Defense Research and Development Canada Valcartier (Quebec)}, 2007.

\bibitem{levesque2009automatic}
M.~Levesque, ``Automatic reacquisition of satellite positions by detecting their expected streaks in astronomical images,'' in \emph{Proceedings of the Advanced Maui Optical and Space Surveillance Technologies Conference}.\hskip 1em plus 0.5em minus 0.4em\relax Citeseer, 2009, p. E81.

\bibitem{jia2020detection}
P.~Jia, Q.~Liu, and Y.~Sun, ``Detection and classification of astronomical targets with deep neural networks in wide-field small aperture telescopes,'' \emph{The Astronomical Journal}, vol. 159, no.~5, p. 212, 2020.

\bibitem{shen2023co}
Z.~Shen, P.~Cao, H.~Yang, X.~Liu, J.~Yang, and O.~R. Zaiane, ``Co-training with high-confidence pseudo labels for semi-supervised medical image segmentation,'' \emph{arXiv preprint arXiv:2301.04465}, 2023.

\bibitem{tarvainen2017mean}
A.~Tarvainen and H.~Valpola, ``Mean teachers are better role models: Weight-averaged consistency targets improve semi-supervised deep learning results,'' \emph{Advances in neural information processing systems}, vol.~30, 2017.

\bibitem{liu2021unbiased}
Y.-C. Liu, C.-Y. Ma, Z.~He, C.-W. Kuo, K.~Chen, P.~Zhang, B.~Wu, Z.~Kira, and P.~Vajda, ``Unbiased teacher for semi-supervised object detection,'' \emph{arXiv preprint arXiv:2102.09480}, 2021.

\bibitem{yang2021interactive}
Q.~Yang, X.~Wei, B.~Wang, X.-S. Hua, and L.~Zhang, ``Interactive self-training with mean teachers for semi-supervised object detection,'' in \emph{Proceedings of the IEEE/CVF conference on computer vision and pattern recognition}, 2021, pp. 5941--5950.

\bibitem{mao2023semi}
Z.~Mao, X.~Tong, and Z.~Luo, ``Semi-supervised remote sensing image change detection using mean teacher model for constructing pseudo-labels,'' in \emph{ICASSP 2023-2023 IEEE International Conference on Acoustics, Speech and Signal Processing (ICASSP)}.\hskip 1em plus 0.5em minus 0.4em\relax IEEE, 2023, pp. 1--5.

\bibitem{yang2022st}
L.~Yang, W.~Zhuo, L.~Qi, Y.~Shi, and Y.~Gao, ``St++: Make self-training work better for semi-supervised semantic segmentation,'' in \emph{Proceedings of the IEEE/CVF conference on computer vision and pattern recognition}, 2022, pp. 4268--4277.

\bibitem{zhu2024sstd}
Z.~Zhu, A.~Zia, X.~Li, B.~Dan, Y.~Ma, E.~Liu, and R.~Zhao, ``Sstd: Stripe-like space target detection using single-point supervision,'' \emph{arXiv preprint arXiv:2407.18097}, 2024.

\bibitem{jiang2022space}
P.~Jiang, C.~Liu, W.~Yang, Z.~Kang, C.~Fan, and Z.~Li, ``Space debris automation detection and extraction based on a wide-field surveillance system,'' \emph{The Astrophysical Journal Supplement Series}, vol. 259, no.~1, p.~4, 2022.

\bibitem{cegarra2022real}
M.~Cegarra~Polo, T.~Yanagisawa, and H.~Kurosaki, ``Real-time processing pipeline for automatic streak detection in astronomical images implemented in a multi-gpu system,'' \emph{Publications of the Astronomical Society of Japan}, vol.~74, no.~4, pp. 777--790, 2022.

\bibitem{nir2018optimal}
G.~Nir, B.~Zackay, and E.~O. Ofek, ``Optimal and efficient streak detection in astronomical images,'' \emph{The Astronomical Journal}, vol. 156, no.~5, p. 229, 2018.

\bibitem{dawson2016blind}
W.~A. Dawson, M.~D. Schneider, and C.~Kamath, ``Blind detection of ultra-faint streaks with a maximum likelihood method,'' \emph{arXiv preprint arXiv:1609.07158}, 2016.

\bibitem{sara2017faint}
R.~Sara and V.~Cvrcek, ``Faint streak detection with certificate by adaptive multi-level bayesian inference,'' in \emph{European Conference on Space Debris}, 2017.

\bibitem{virtanen2016streak}
J.~Virtanen, J.~Poikonen, T.~S{\"a}ntti, T.~Komulainen, J.~Torppa, M.~Granvik, K.~Muinonen, H.~Pentik{\"a}inen, J.~Martikainen, J.~N{\"a}r{\"a}nen \emph{et~al.}, ``Streak detection and analysis pipeline for space-debris optical images,'' \emph{Advances in Space Research}, vol.~57, no.~8, pp. 1607--1623, 2016.

\bibitem{huang1979fast}
T.~Huang, G.~Yang, and G.~Tang, ``A fast two-dimensional median filtering algorithm,'' \emph{IEEE transactions on acoustics, speech, and signal processing}, vol.~27, no.~1, pp. 13--18, 1979.

\bibitem{serra1992overview}
J.~Serra and L.~Vincent, ``An overview of morphological filtering,'' \emph{Circuits, Systems and Signal Processing}, vol.~11, pp. 47--108, 1992.

\bibitem{xi2016space}
J.~Xi, D.~Wen, O.~K. Ersoy, H.~Yi, D.~Yao, Z.~Song, and S.~Xi, ``Space debris detection in optical image sequences,'' \emph{Applied optics}, vol.~55, no.~28, pp. 7929--7940, 2016.

\bibitem{duarte2023space}
P.~Duarte, P.~Gordo, N.~Peixinho, R.~Melicio, D.~Val{\'e}rio, R.~Gafeira \emph{et~al.}, ``Space surveillance payload camera breadboard: Star tracking and debris detection algorithms,'' \emph{Advances in Space Research}, vol.~72, no.~10, pp. 4215--4228, 2023.

\bibitem{UIU-Net}
X.~Wu, D.~Hong, and J.~Chanussot, ``Uiu-net: U-net in u-net for infrared small object detection,'' \emph{IEEE Transactions on Image Processing}, vol.~32, pp. 364--376, 2022.

\bibitem{DNANET}
B.~Li, C.~Xiao, L.~Wang, Y.~Wang, Z.~Lin, M.~Li, W.~An, and Y.~Guo, ``Dense nested attention network for infrared small target detection,'' \emph{IEEE Transactions on Image Processing}, vol.~32, pp. 1745--1758, 2022.

\bibitem{ACM}
Y.~Dai, Y.~Wu, F.~Zhou, and K.~Barnard, ``Asymmetric contextual modulation for infrared small target detection,'' in \emph{Proceedings of the IEEE/CVF Winter Conference on Applications of Computer Vision}, 2021, pp. 950--959.

\bibitem{RDIAN}
H.~Sun, J.~Bai, F.~Yang, and X.~Bai, ``Receptive-field and direction induced attention network for infrared dim small target detection with a large-scale dataset irdst,'' \emph{IEEE Transactions on Geoscience and Remote Sensing}, vol.~61, pp. 1--13, 2023.

\bibitem{ISNET}
M.~Zhang, R.~Zhang, Y.~Yang, H.~Bai, J.~Zhang, and J.~Guo, ``Isnet: Shape matters for infrared small target detection,'' in \emph{Proceedings of the IEEE/CVF Conference on Computer Vision and Pattern Recognition}, 2022, pp. 877--886.

\bibitem{AGPCNET}
T.~Zhang, S.~Cao, T.~Pu, and Z.~Peng, ``Agpcnet: Attention-guided pyramid context networks for infrared small target detection,'' \emph{arXiv preprint arXiv:2111.03580}, 2021.

\bibitem{lu2023mutually}
S.~Lu, Z.~Zhang, Z.~Yan, Y.~Wang, T.~Cheng, R.~Zhou, and G.~Yang, ``Mutually aided uncertainty incorporated dual consistency regularization with pseudo label for semi-supervised medical image segmentation,'' \emph{Neurocomputing}, vol. 548, p. 126411, 2023.

\bibitem{wu2023compete}
H.~Wu, X.~Li, Y.~Lin, and K.-T. Cheng, ``Compete to win: Enhancing pseudo labels for barely-supervised medical image segmentation,'' \emph{IEEE Transactions on Medical Imaging}, vol.~42, no.~11, pp. 3244--3255, 2023.

\bibitem{chen2023semiroadexnet}
H.~Chen, Z.~Li, J.~Wu, W.~Xiong, and C.~Du, ``Semiroadexnet: A semi-supervised network for road extraction from remote sensing imagery via adversarial learning,'' \emph{ISPRS Journal of Photogrammetry and Remote Sensing}, vol. 198, pp. 169--183, 2023.

\bibitem{han2024c2f}
C.~Han, C.~Wu, M.~Hu, J.~Li, and H.~Chen, ``C2f-semicd: A coarse-to-fine semi-supervised change detection method based on consistency regularization in high-resolution remote-sensing images,'' \emph{IEEE Transactions on Geoscience and Remote Sensing}, 2024.

\bibitem{wang2021semi}
W.~Wang and C.~Su, ``Semi-supervised semantic segmentation network for surface crack detection,'' \emph{Automation in Construction}, vol. 128, p. 103786, 2021.

\bibitem{jian2024cross}
Z.~Jian and J.~Liu, ``Cross teacher pseudo supervision: Enhancing semi-supervised crack segmentation with consistency learning,'' \emph{Advanced Engineering Informatics}, vol.~59, p. 102279, 2024.

\bibitem{sajjadi2016regularization}
M.~Sajjadi, M.~Javanmardi, and T.~Tasdizen, ``Regularization with stochastic transformations and perturbations for deep semi-supervised learning,'' \emph{Advances in neural information processing systems}, vol.~29, 2016.

\bibitem{laine2016temporal}
S.~Laine and T.~Aila, ``Temporal ensembling for semi-supervised learning,'' \emph{arXiv preprint arXiv:1610.02242}, 2016.

\bibitem{french2019semi}
G.~French, T.~Aila, S.~Laine, M.~Mackiewicz, and G.~Finlayson, ``Semi-supervised semantic segmentation needs strong, high-dimensional perturbations,'' 2019.

\bibitem{lee2013pseudo}
D.-H. Lee \emph{et~al.}, ``Pseudo-label: The simple and efficient semi-supervised learning method for deep neural networks,'' in \emph{Workshop on challenges in representation learning, ICML}, vol.~3, no.~2.\hskip 1em plus 0.5em minus 0.4em\relax Atlanta, 2013, p. 896.

\bibitem{sohn2020simple}
K.~Sohn, Z.~Zhang, C.-L. Li, H.~Zhang, C.-Y. Lee, and T.~Pfister, ``A simple semi-supervised learning framework for object detection,'' \emph{arXiv preprint arXiv:2005.04757}, 2020.

\bibitem{xu2021end}
M.~Xu, Z.~Zhang, H.~Hu, J.~Wang, L.~Wang, F.~Wei, X.~Bai, and Z.~Liu, ``End-to-end semi-supervised object detection with soft teacher,'' in \emph{Proceedings of the IEEE/CVF international conference on computer vision}, 2021, pp. 3060--3069.

\bibitem{zhou2023semi}
Y.~Zhou, X.~Jiang, Z.~Chen, L.~Chen, and X.~Liu, ``A semi-supervised arbitrary-oriented sar ship detection network based on interference consistency learning and pseudo label calibration,'' \emph{IEEE Journal of Selected Topics in Applied Earth Observations and Remote Sensing}, 2023.

\bibitem{wang2022uctransnet}
H.~Wang, P.~Cao, J.~Wang, and O.~R. Zaiane, ``Uctransnet: rethinking the skip connections in u-net from a channel-wise perspective with transformer,'' in \emph{Proceedings of the AAAI conference on artificial intelligence}, vol.~36, no.~3, 2022, pp. 2441--2449.

\bibitem{ma2023msma}
T.~Ma, H.~Wang, J.~Liang, J.~Peng, Q.~Ma, and Z.~Kai, ``Msma-net: An infrared small target detection network by multi-scale super-resolution enhancement and multi-level attention fusion,'' \emph{IEEE Transactions on Geoscience and Remote Sensing}, 2023.

\bibitem{ronneberger2015u}
O.~Ronneberger, P.~Fischer, and T.~Brox, ``U-net: Convolutional networks for biomedical image segmentation,'' in \emph{Medical image computing and computer-assisted intervention--MICCAI 2015: 18th international conference, Munich, Germany, October 5-9, 2015, proceedings, part III 18}.\hskip 1em plus 0.5em minus 0.4em\relax Springer, 2015, pp. 234--241.

\end{thebibliography}
\end{document}